\documentclass[sigconf]{acmart}

\usepackage{booktabs}
\usepackage{multirow}
\usepackage{threeparttable}
\usepackage{tabularx}
\usepackage{subcaption}
\usepackage{xcolor}
\usepackage{xparse}
\usepackage{xspace}
\usepackage{pifont}
\usepackage{enumitem}
\usepackage{noindentafter}
\usepackage[noindentafter]{titlesec}

\AtBeginDocument{%
  \providecommand\BibTeX{{%
    \normalfont B\kern-0.5em{\scshape i\kern-0.25em b}\kern-0.8em\TeX}}}

\newcommand{\name}{\tool{Type4Py}}

\NewDocumentCommand\code{+m}{\texttt{#1}}
\NewDocumentCommand\tool{+m}{{\small\scshape #1}\xspace}

\newcommand{\cmark}{\ding{51}}
\newcommand{\xmark}{\ding{55}}

\NoIndentAfterEnv{itemize}
\NoIndentAfterEnv{description}
\NoIndentAfterEnv{enumerate}
\setlist{
	leftmargin=12pt,
	itemsep=1pt plus 1pt,
	topsep=2pt plus 1pt,
	parsep=2pt plus 1pt,
	listparindent=0pt,
}
\setlist[description]{
	font=\normalfont\itshape}

\titleformat{\subsubsection}[runin]{\itshape}{\arabic{subsubsection}) }{0pt}{}[:] 
\titleformat{\paragraph}[runin]{\itshape}{}{0pt}{}[:] 
\titlespacing{\subsubsection}{0pt}		{\smallskipamount}{4pt}
\titlespacing{\paragraph}{0pt}		{\smallskipamount}{4pt}
\titlespacing{\subsection}{0pt}		{\smallskipamount}{2pt}
\titlespacing{\section}{0pt}		{\smallskipamount}{2pt}

\copyrightyear{2022} 
\acmYear{2022} 
\setcopyright{rightsretained} 
\acmConference[ICSE '22]{44th International Conference on Software Engineering}{May 21--29, 2022}{Pittsburgh, PA, USA}
\acmBooktitle{44th International Conference on Software Engineering (ICSE '22), May 21--29, 2022, Pittsburgh, PA, USA}
\acmDOI{10.1145/3510003.3510124}
\acmISBN{978-1-4503-9221-1/22/05}

\begin{document}

\title{Type4Py: Practical Deep Similarity Learning-Based Type Inference for Python}

\author{Amir M. Mir}
\email{s.a.m.mir@tudelft.nl}
\affiliation{%
  \institution{Delft University of Technology}
  \city{Delft}
  \country{The Netherlands}
}

\author{Evaldas Latoškinas}
\email{e.latoskinas@student.tudelft.nl}
\affiliation{%
	\institution{Delft University of Technology}
	\city{Delft}
	\country{The Netherlands}
}

\author{Sebastian Proksch}
\email{s.proksch@tudelft.nl}
\affiliation{%
	\institution{Delft University of Technology}
	\city{Delft}
	\country{The Netherlands}
}

\author{Georgios Gousios}
\email{gousiosg@fb.com}
\affiliation{%
  \institution{Facebook}
  \city{Menlo Park}
  \country{USA}}

\renewcommand{\shortauthors}{Mir et al.}

\begin{abstract}
Dynamic languages, such as Python and Javascript, trade static typing for 
developer flexibility and productivity. Lack of static typing can cause run-time exceptions and is a
major factor for weak IDE support. To alleviate these issues, PEP 484 introduced
optional type annotations for Python. As retrofitting types to existing
codebases is error-prone and laborious, machine learning (ML)-based approaches have been proposed to
enable automatic type inference based on existing, partially annotated
codebases.
However, previous ML-based approaches are trained and evaluated on human-provided type annotations, 
which might not always be sound, and hence this may limit the practicality for real-world usage.
In this paper, we present \name, a deep similarity learning-based hierarchical neural network model.
It learns to discriminate between similar and dissimilar types in a high-dimensional space, which results in clusters of types.
Likely types for arguments, variables, and return values can then be inferred through the nearest neighbor search.
Unlike previous work, we trained and evaluated our model on a \emph{type-checked} dataset and used mean reciprocal rank (MRR) to reflect the performance perceived by users.
The obtained results show that \name achieves an MRR of 77.1\%, which is a substantial improvement of 8.1\% and 16.7\% over the state-of-the-art approaches \tool{Typilus} and \tool{TypeWriter}, respectively. Finally, to aid developers with retrofitting types, we released a Visual Studio Code extension, which uses \name to provide ML-based type auto-completion for Python.
\end{abstract}





\maketitle

\section{Introduction}

Over the past years, \emph{dynamically-typed} programming languages (DPLs) have become extremely popular among software developers.
The IEEE Spectrum ranks Python as the most popular programming language in 2021~\cite{ieeespec2019}.
It is known that \emph{statically-typed} languages are less error-prone~\cite{ray2014large} and that static types improve important quality aspects of software~\cite{gao2017type}, like the maintainability of software systems in terms of understandability, fixing type errors~\cite{hanenberg2014empirical}, and early bug detection~\cite{gao2017type}.
In contrast to that, dynamic languages such as Python and JavaScript allow rapid prototyping which potentially reduces development time~\cite{hanenberg2014empirical, stuchlik2011static}, but the lack of static types in dynamically-typed languages often leads to type errors, unexpected run-time behavior, and suboptimal IDE support.

To mitigate these shortcomings, the Python community introduced \emph{PEP 484}~\cite{van2014pep}, which adds optional static typing to Python 3.5 and newer.
Static type inference methods~\cite{hassan2018maxsmt, furr2009static} can be employed to support adding these annotations, which is otherwise a manual, cumbersome, and error-prone process~\cite{ore2018assessing}.
However, static inference is imprecise~\cite{pavlinovic2019leveraging}, caused by dynamic language features or by the required over-approximation of program behavior~\cite{madsen2015static}.
Moreover, static analysis is usually performed on full programs, including their dependencies, which is slow and resource-intensive.

To address these limitations of static type inference methods, researchers have recently employed \emph{Machine Learning} (ML) techniques for type prediction in dynamic languages~\cite{hellendoorn2018deep, malik2019nl2type, pradel2019typewriter, allamanis2020typilus}.
The experimental results of these studies show that ML-based type prediction approaches are more precise than static type inference methods or they can also work with static methods in a complementary fashion~\cite{pradel2019typewriter, allamanis2020typilus}. Despite the superiority of ML-based type prediction approaches, their type vocabulary is small and fixed-sized (i.e. 1,000 types). This limits their type prediction ability for user-defined and rare types. To solve this issue, Allamanis et al.~\cite{allamanis2020typilus} recently introduced \tool{Typilus} that does not constraint the type vocabulary size and it outperforms the other models with small-sized type vocabulary.

While the ML-based type inference approaches are effective, we believe that there are two main drawbacks in the recent previous work~\cite{pradel2019typewriter, allamanis2020typilus}:
\begin{itemize}
	\item The neural models are trained and evaluated on developer-provided type annotations, which are not always correct~\cite{ore2018assessing, rak2020python}. This might be a (major) threat to the validity of the obtained results. To address this, a type checker should be employed to detect and remove incorrect type annotations from the dataset.
	\item Although the proposed approaches~\cite{pradel2019typewriter, allamanis2020typilus} obtain satisfying performance for Top-10, it is important for an approach to give a correct prediction in Top-1 as developers tend to use the first suggestion by a tool~\cite{parnin2011automated}. Like the API recommendation research~\cite{liu2018effective, he2021pyart}, the Mean Reciprocal Rank (MRR) metric should also be used for evaluation, which \emph{paratially} rewards an approach where the correct API is not in the Top-1 suggestion.
\end{itemize}

Motivated by the above discussion, we present \name, a type inference approach based on \emph{deep similarity learning} (DSL).
The proposed approach consists of an effective hierarchical neural network that maps programs into \emph{type clusters} in a high-dimensional feature space.
Similarity learning has, for example, been used in Computer Vision to discriminate human faces for verification~\cite{chopra2005learning}. Similarly, \name learns how to distinguish between different types through a DSL-based hierarchical neural network.
As a result, our proposed approach can not only handle a very large type vocabulary, but also it can be used in practice by developers for retrofitting type annotations.
We compare to the state-of-the art approaches and the experimental results show that \name obtains an MRR of 77.1\%,
which is 8.1\% and 16.7\% higher than \tool{Typilus}~\cite{allamanis2020typilus} and \tool{TypeWriter}~\cite{pradel2019typewriter}, respectively.

\smallskip
\noindent
Overall, this paper presents the following main contributions:
\begin{itemize}
	\item \name, a new DSL-based type inference approach.
	\item A \emph{type-checked} dataset with 5.1K Python projects and 1.2M type annotations. Invalid type annotations are removed from both training and evaluation.
	\item A Visual Studio Code extension~\cite{vscodet4py}, which provides ML-based type auto-completion for Python.
\end{itemize}

To foster future research, we publicly released the implementation of the \name model and its dataset on GitHub.\footnote{https://github.com/saltudelft/type4py}

The rest of the paper is organized as follows. Section \ref{sec:rw} reviews related work on static and ML-based type inference. The proposed approach, \name, is described in Section \ref{sec:pa}. Section \ref{sec:data} gives details about the creation of the type-checked dataset for evaluation. The evaluation setup and empirical results are given in Section \ref{sec:es} and Section \ref{sec:eval}, respectively. Section \ref{sec:prac} describes the deployment of \name and its usage in Visual Studio Code. Section \ref{sec:discuss} discusses the obtained results and gives future directions. Finally, we summarize our work in Section \ref{sec:sum}.

\section{Related Work}\label{sec:rw}
\begin{table*}[!t]
	\centering
	\caption{Comparison between \name and other learning-based type inference approaches}
	\label{tab:comp-learning-appr}
	\begin{threeparttable}
	\begin{tabular}{c c c c c c c c c c}
		\toprule
		\multirow{2}{*}{Approach} & \multirow{2}{*}{Size of type vocabulary} & \multirow{2}{*}{ML model} & \multicolumn{3}{c}{Type hints} & &  \multicolumn{3}{c}{Supported Predictions} \\
		\cmidrule{4-6} \cmidrule{8-10}
		& & & Contexual & Natural & Logical & & Argument & Return & Variable \\
		\midrule
		\textbf{\name} & Unlimited & HNN (2x RNNs) & \cmark & \cmark & \xmark & & \cmark & \cmark & \cmark \\
		JSNice \cite{raychev2015predicting} & 10+ & CRFs & \cmark & \cmark & \xmark &&  \cmark & \xmark & \xmark \\
		Xu et al. \cite{xu2016python} & - & PGM & \xmark & \cmark & \cmark & & \xmark & \xmark & \cmark \\
		DeepTyper \cite{hellendoorn2018deep} & 10K+ & biRNN & \cmark & \cmark & \xmark & & \cmark & \cmark & \cmark \\
		NL2Type \cite{malik2019nl2type} & 1K & LSTM & \xmark & \cmark & \xmark &  & \cmark & \cmark & \xmark \\
		TypeWriter \cite{pradel2019typewriter} & 1K & HNN (3x RNNs) & \cmark & \cmark & \xmark & & \cmark & \cmark & \xmark  \\
		LAMBDANET \cite{wei2019lambdanet} & 100\tnote{a} & GNN & \cmark & \cmark & \cmark & & \xmark & \xmark & \cmark \\
		OptTyper \cite{pandi2020opttyper} & 100 & LSTM & \xmark & \cmark & \cmark &  & \cmark & \cmark & \xmark \\
		Typilus \cite{allamanis2020typilus} & Unlimited & GNN & \cmark & \cmark & \xmark  & & \cmark & \cmark & \cmark \\
		TypeBert \cite{jesse2021learning} & 40K & BERT & \cmark & \cmark & \xmark & & \cmark & \cmark & \cmark \\
		\bottomrule
	\end{tabular}
    \begin{tablenotes}
    	\item[a] {\footnotesize Note that LAMBDANET's pointer network model enables to predict user-defined types outside its fixed-size type vocabulary.}
    \end{tablenotes}
    \end{threeparttable}
\end{table*}
\paragraph{Type checking  and inference for Python}
In 2014, the Python community introduced a type hints proposal \cite{van2014pep} that describes adding optional type annotations to Python programs. A year later, Python 3.5 was released with optional type annotations and the \textit{mypy} type checker \cite{lehtosalo2017mypy}. This has enabled gradual typing of existing Python programs and validating added type annotations. Since the introduction of type hints proposal, other type checkers have been developed such as \textit{PyType} \cite{pytype}, \textit{PyRight} \cite{pyright}, and \textit{Pyre} \cite{pyre}.

A number of research works proposed type inference algorithms for Python \cite{salib2004faster, maia2012static, hassan2018maxsmt}. These are static-based approaches that have a pre-defined set of rules and constraints. As previously mentioned, static type inference methods are often imprecise \cite{pavlinovic2019leveraging}, due to the dynamic nature of Python and the over-approximation of programs' behavior by static analysis \cite{madsen2015static}.

\paragraph{Learning-based type inference} In 2015, Rachev et al.~\cite{raychev2015predicting} proposed JSNice, a probabilistic model that predicts identifier names and type annotations for JavaScript using conditional random fields (CRFs). The central idea of JSNice is to capture relationships between program elements in a dependency network.
However, the main issue with JSNice is that its dependency network cannot consider a wide context within a program or a function.

Xu et al.~\cite{xu2016python} adopt a probabilistic graphical model (PGM) to predict variable types for Python. Their approach extracts several uncertain type hints such as attribute access, variable names, and data flow between variables. Although the probabilistic model of Xu et al.~\cite{xu2016python} outperforms static type inference systems, their proposed system is slow and lacks scalability.

Considering the mentioned issue of JSNice, Hellendoorn et al.~\cite{hellendoorn2018deep} proposed DeepTyper, a sequence-to-sequence neural network model that was trained on an aligned corpus of TypeScript code. The DeepTyper model can predict type annotations across a source code file by considering a much wider context. Yet DeepTyper suffers from inconsistent predictions for the token-level occurrences of the same variable. Malik et al.~\cite{malik2019nl2type} proposed NL2Type, a neural network model that predicts type annotations for JavaScript functions. The basic idea of NL2Type is to leverage the natural language information in the source code such as identifier names and comments. The NL2Type model is shown to outperform both the JSNice and DeepTyper at the task of type annotations prediction~\cite{malik2019nl2type}.

Motivated by the NL2Type model, Pradel et al.~\cite{pradel2019typewriter} proposed the TypeWriter model which infers type annotations for Python. TypeWriter is a deep neural network model that considers both code context and natural language information in the source code. Moreover, TypeWriter validates its neural model's type predictions by employing a combinatorial search strategy and an external type checker. Wei et al.~\cite{wei2019lambdanet} introduced LAMBDANET, a graph neural network-based type inference for TypeScript. Its main idea is to create a type dependency graph that links to-be-typed variables with logical constraints and contextual hints such as variables assignments and names. For type prediction, LAMBDANET employs a pointer-network-like model which enables the prediction of unseen user-defined types. The experimental results of Wei et al.~\cite{wei2019lambdanet} show the superiority of LAMBDANET over DeepTyper.

Given that the natural constraints such as identifiers and comments are an uncertain source of information, Pandi et al.~\cite{pandi2020opttyper} proposed OptTyper which predicts types for the TypeScript language. The central idea of their approach is to extract deterministic information or logical constraints from a type system and combine them with the natural constraints in a single optimization problem. This allows OptTyper to make a type-correct prediction without violating the typing rules of the language. OptTyper has been shown to outperform both LAMBDANET and DeepTyper~\cite{pandi2020opttyper}.

Except for LAMBDANET, all the discussed learning-based type inference methods employ a (small) fixed-size type vocabulary, e.g., 1,000 types. This hinders their ability to infer user-defined and rare types. To address this, Allamanis et al.~\cite{allamanis2020typilus} proposed Typilus, which is a graph neural network (GNN)-based model that integrates information from several sources such as identifiers, syntactic patterns, and data flow to infer type annotations for Python. Typilus is based on metric-based learning and learns to discriminate similar to-be-typed symbols from different ones. However, Typilus requires a sophisticated source code analysis to create its graph representations, i.e. data flow analysis. Very recently, inspired by "Big Data", Jesse et al. ~\cite{jesse2021learning} presented TypeBert, a pre-trained BERT model with simple token-sequence representation. Their empirical results show that TypeBert generally outperforms LAMBDANET. The differences between \name and other learning-based approaches are summarized in Table \ref{tab:comp-learning-appr}.

\section{Proposed Approach}\label{sec:pa}

\begin{figure*}[!t]
	\centering
	\includegraphics[width=\linewidth]{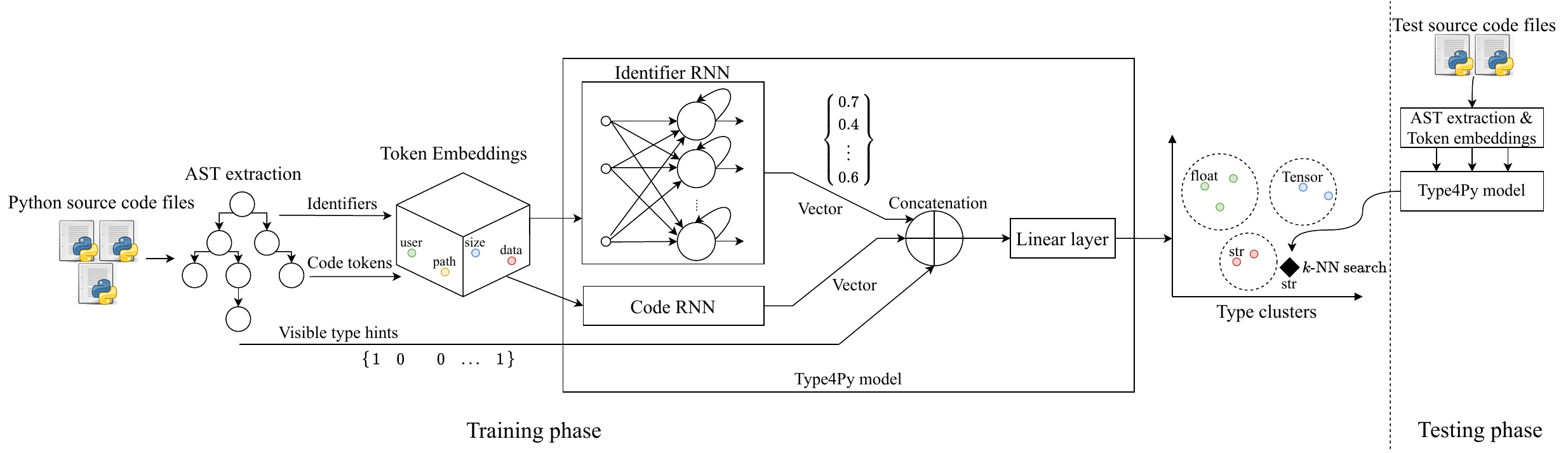}
	\caption{Overview of \name approach}
	\label{fig:overview-approach}
\end{figure*}

This section presents the details of \name by going through the different steps of the pipeline that is illustrated in the overview of the proposed approach in Figure~\ref{fig:overview-approach}.
We first describe how we extract type hints from Python source code and then how we use this information to train the neural model.

\subsection{Type hints}
We extract the Abstract Syntax Tree (AST) from Python source code files. By traversing the nodes of ASTs, we obtain type hints that are valuable for predicting types of function arguments, variables, and return types. The obtained type hints are based on natural information, code context, and import statements which are described in this section.

\paragraph{Natural Information}
As indicated by the previous work~\cite{hindle2012naturalness, malik2019nl2type}, source code contains useful and informal natural language information that is considered as a source of type hints. In DPLs, developers tend to name variables and functions' arguments after their expected type~\cite{milojkovic2017exploiting}. Based on this intuition, we consider identifier names as a main source of natural information and type hint. Specifically, we extract the name of functions ($N_{f}$) and their arguments ($N_{args}$) as they may provide a hint about the return type of functions and the type of functions' arguments, respectively. We also denote a function's argument as $N_{arg}$ hereafter. For variables, we extract their names as denoted by $N_{v}$.


\paragraph{Code Context}
We extract all uses of an argument in the function body as a type hint.
This means that the complete statement, in which the argument is used, is included as a sequence of tokens. Similarly, we extract all uses of a variable in its current and inner scopes.
Also, all the return statements inside a function are extracted as they may contain a hint about the return type of the function.

\paragraph{Visible type hints (VTH)}
In contrast to previous work that only analyzed the direct imports~\cite{pradel2019typewriter}, we recursively extract all the import statements in a given module and its transitive dependencies.
We build a dependency graph for all imports of user-defined classes, type aliases, and \texttt{NewType} declarations
For example, if module \texttt{A} imports \texttt{B.Type} and \texttt{C.D.E}, the edges (\texttt{A}, \texttt{B.Type}) and (\texttt{A}, \texttt{C.D.E}) will be added to the graph.
We expand wildcard imports like \texttt{from foo import *} and resolve the concrete type references.
We consider the identified types as \emph{visible} and store them with their fully-qualified name to reduce ambiguity. For instance, \texttt{tf.Tensor} and \texttt{torch.Tensor} are different types.
Although the described inspection-based approach is slower than a pure AST-based analysis, our ablation analysis shows that VTHs substantially improve the performance of \name (subsection~\ref{subsec:abalation}).

\subsection{Vector Representation}
In order for a machine learning model to learn from type hints, they are represented as real-valued vectors. The vectors preserve semantic similarities between similar words. To capture those, a word embedding technique is used to map words into a $d$-dimensional vector space, $\mathbb{R}^{d}$. Specifically, we first preprocess extracted identifiers and code contexts by applying common Natural Language Processing (NLP) techniques. This preprocessing step involves tokenization, stop word removal, and lemmatization~\cite{JurafskyNLP}. Afterwards, we employ Word2Vec \cite{mikolov2013distributed} embeddings to train a code embedding $E_{c}: w_{1},\dots,w_{l} \to \mathbb{R}^{l \times d}$ for both code context and identifier tokens, where $w_{i}$ and $l$ denote a single token and the length of a sequence, respectively. In the following, we describe the vector representation of all the three described type hints for both argument types and return types.

\paragraph{Identifiers} Given an argument's type hints, the vector sequence of the argument is represented as follows:
\begin{equation*}
E_{c}(N_{arg}) \, \circ s \, \circ E_{c}(N_{f}) \, \circ E_{c}(N_{args})
\end{equation*}
where $\circ$ concatenates and flattens sequences, and $s$ is a separator\footnote{The separator is a vector of ones with appropriate dimension.}. For a return type, its vector sequence is represented as follows:
\begin{equation*}
E_{c}(N_{f}) \, \circ s \, \circ E_{c}(N_{args})
\end{equation*}
Last, a variable's identifier is embedded as $E_{c}({N_{v}})$.

\paragraph{Code contexts} For function arguments and variables, we concatenate the sequences of their usages into a single sequence. Similarly, for return types, we concatenate all the return statements of a function into a single sequence. To truncate long sequences, we consider a window of $n$ tokens at the center of the sequence (default $n=7$). Similar to identifiers, the function embedding $E_{c}$ is used to convert code contexts sequences into a real-valued vector.

\paragraph{Visible type hints} Given all the source code files, we build a fixed-size vocabulary of visible type hints. The vocabulary covers the majority of all visible type occurrences. Because most imported visible types in Python modules are built-in primitive types such as \texttt{List}, \texttt{Dict}, and their combinations. If a type is out of the visible type vocabulary, it is represented as a special \texttt{other} type. For function arguments, variables, and return types, we create a sparse binary vector of size $T$ whose elements represent a type. An element of the binary vector is set to one if and only if its type is present in the vocabulary. Otherwise, the \texttt{other} type is set to one in the binary vector.

\subsection{Neural model}
The neural model of our proposed approach employs a hierarchical neural network (HNN), which consists of two recurrent neural networks (RNNs)~\cite{williams1989learning}. HNNs are well-studied and quite effective for text and vision-related tasks~\cite{liu2020hienn, zheng2019hierarchical, du2015hierarchical}. In the case of type prediction, intuitively, HNNs can capture different aspects of identifiers and code context. In the neural architecture (see Fig. \ref{fig:overview-approach}), the two RNNs are based on long short-term memory (LSTM) units~\cite{hochreiter1997long}. Here, we chose LSTMs units as they are effective for capturing long-range dependencies~\cite{goodfellow2016deep}. Also, LSTM-based neural models have been applied successfully to NLP tasks such as sentiment classification~\cite{rao2018lstm}. Formally, the output $h_{i}^{(t)}$ of the $i$-th LSTM unit at the time step $t$ is defined as follows:
\begin{equation}
h_{i}^{(t)} = \tanh(s_{i}^{t}) \, \sigma\left( b_{i} +  \sum\limits_{j}{U_{i,j}x_{j}^{(t)} + \sum\limits_{j}{W_{i,j}h_{j}^{(t-1)}}} \right)
\end{equation}

\noindent which has sigmoid function $\sigma$, current input vector $x_{j}$, unit state $s_{i}^{t}$ and has model parameters $W$, $U$, $b$ for its recurrent weights, input weights and biases~\cite{goodfellow2016deep}. The two hierarchical RNNs allow capturing different aspects of input sequences from identifiers and code tokens. The captured information is then summarized into two single vectors, which are obtained from the final hidden state of their corresponding RNN. The two single vectors from RNNs are concatenated with the visible type hints vector and the resulting vector is passed through a fully-connected linear layer.

In previous work~\cite{pradel2019typewriter, malik2019nl2type}, the type prediction task is formulated as a classification problem. As a result, the linear layer of their neural model outputs a vector of size 1,000 with probabilities over predicted types. Therefore, the neural model predicts \textit{unkonwn} if it has not seen a type in the training phase. To address this issue, we formulate the type prediction task as a Deep Similarity Learning problem~\cite{chopra2005learning, liao2017triplet}. By using the DSL formulation, our neural model learns to map argument, variable, return types into real continuous space, called \textit{type clusters} (also known as type space in~\cite{allamanis2020typilus}). In other words, our neural model maps similar types (e.g. \texttt{str}) into its own type cluster, which should be as far as possible from other clusters of types. Unlike the previous work~\cite{pradel2019typewriter, malik2019nl2type}, our proposed model can handle a very large type vocabulary.

To create the described type clusters, we use \textit{Triplet loss}~\cite{cheng2016person} function which is recently used for computer vision tasks such as face recognition~\cite{cheng2016person}. By using the Triplet loss, a neural model learns to discriminate between similar samples and dissimilar samples by mapping samples into their own clusters in the continuous space. In the case of type prediction, the loss function accepts a type $t_{a}$, a type $t_{p}$ same as $t_{a}$, and a type $t_{n}$ which is different than $t_{a}$. Given a positive scalar margin $m$, the Triplet loss function is defined as follows:
\begin{equation}\label{eq:triplet}
L(t_{a}, t_{p}, t_{n}) = max(0, m + \left\| t_{a} - t_{p} \right\| - \left\| t_{a} - t_{n} \right\|)
\end{equation}

The goal of the objective function $L$ is to make $t_{a}$ examples closer to the similar examples $t_{p}$ than to $t_{n}$ examples. We use Euclidean metric to measure the distance of $t_{a}$ with $t_{p}$ and $t_{n}$.

At prediction time, we first map a query example $t_{q}$ to the type clusters. The query example $t_{q}$ can be a function's argument, the return type of a function or a variable. Then we find the $k$-neareast neighbor (KNN)~\cite{cover1967nearest} of the query example $t_{q}$. Given the $k$-nearest examples $t_{i}$ with a distance $d_{i}$ from the query example $t_{q}$, the probability of $t_{q}$ having a type $t^{\prime}$ can be obtained as follows:
\begin{equation}
P(t_{q}: t^{\prime}) = \frac{1}{N} \sum\limits_{i}^{k}{\frac{\mathbb{I}(t_{i} = t^{\prime} )}{(d_{i} + \varepsilon)^{2} }}
\end{equation}

where $\mathbb{I}$ is the indicator function, $N$ is a normalizing constant, and $\varepsilon$ is a small scalar (i.e. $\varepsilon = 10^{-10}$).

\section{Dataset}\label{sec:data}
For this work, we have created a new version of our ManyTypes4Py dataset~\cite{mt4py2021}, i.e., v0.7. The rest of this section describes the creation of the dataset. To find Python projects with type annotations, on Libraries.io, we searched for projects that depend on the \texttt{mypy} package~\cite{mypy}, i.e., the official and most popular type checker for Python. Intuitively, these projects are more likely to have type annotations. The search resulted in 5.2K Python projects that are available on GitHub. Initially, the dataset has 685K source files and 869K type annotations.

\subsection{Code de-duplication}
On GitHub, Python projects often have file-level duplicates~\cite{lopes2017dejavu} and also code duplication has a negative effect on the performance of ML models when evaluating them on unseen code samples~\cite{allamanis2018adverse}. Therefore, to de-duplicate the dataset, we use our code de-duplication tool, CD4Py~\cite{cd4py}. It uses term frequency-inverse document (TF-IDF) \cite{manning2008introduction} to represent a source code file as a vector in $\mathbb{R}^{n}$ and employ KNN search to find clusters of similar duplicate files. While assuming that the similarity is transitive \cite{allamanis2018adverse}, we keep a file from each cluster and remove all other identified duplicate files from the dataset. Using the described method, we removed around 400K duplicate files from the dataset.

\subsection{Augmentation}
Similar to the work of Allamanis et al.~\cite{allamanis2020typilus}, we have employed a static type inference tool, namely, Pyre~\cite{pyre} v0.9.0 to augment our initial dataset with more type annotations. However, we do note that we could only infer the type of variables using Pyre's \code{query} command. In our experience, the query command could not infer the type of arguments and return types. The command accepts a list of files and returns JSON files containing type information.

\begin{figure}[!t]
	\centering
	\includegraphics[width=\linewidth]{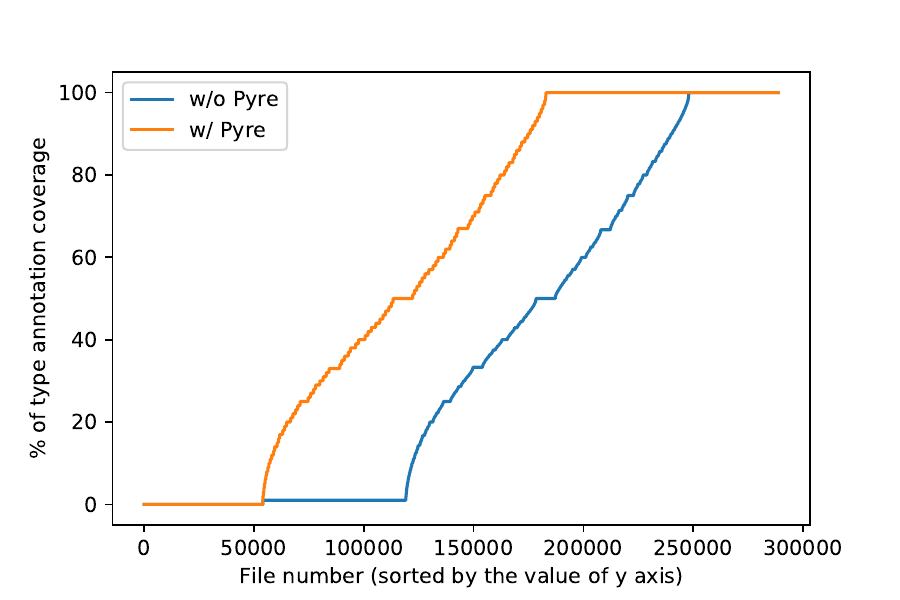}
	\caption{The effect of using Pyre on the type annotation coverage of source code files}
	\label{fig:dataset-type-annot-cove-pyre}
\end{figure}

Thanks to Pyre's inferred types, the dataset has now 3.3M type annotations in total. To demonstrate the effect of using Pyre on the dataset, Figure \ref{fig:dataset-type-annot-cove-pyre} shows the percentage of type annotation coverage for source code files with/without using Pyre. After using Pyre, of 288,760 source code files, 65\% of them have more than 40\% type annotation coverage.

\subsection{Type Checking}
Recent studies show that developer-provided types rarely type-check and Python projects may contain type-related defects~\cite{ore2018assessing, rak2020python, khan2021empirical}. Therefore, we believe that it is essential to type-check the dataset to eliminate noisy ground truth (i.e. incorrect type annotations). Not only noisy ground truth can be considered a threat to the validity of results but also it may make the discrimination of types in type clusters more difficult~\cite{garcia2015effect}. To clean the dataset from noisy ground truth, we perform basic analysis as follows:
\begin{itemize}
	\item First, we use mypy to type-check 288,760 source files in the dataset. Of which, 184,752 source files are successfully type-checked.
	\item Considering the remaining 104,008 source files, for further analysis, we ignore source files that cannot be type-checked further by mypy due to the syntax error or other fatal exceptions. This amounts to 63,735 source files in the dataset.
	\item Given 40,273 source files with type errors, we remove one type annotation at a time from a file and run mypy. If it type-checks, we include the file. Otherwise, we continue this step up to 10 times. This basic analysis fixes 16,861 source files with type errors, i.e, 42\% of the given set of files.
\end{itemize}

\begin{table}[!t]
	\centering
	\caption{Characteristics of the dataset used for evaluation}
	\label{tab:dataset}
	\begin{threeparttable}
		\begin{tabularx}{\linewidth}{@{}l l @{}}
			\toprule
			Metrics\tnote{a,b} & Our dataset \\
			\midrule
			Repositories & 5,092 \\
			Files & 201,613 \\
			Lines of code\tnote{c} & 11.9M \\
			\midrule
			Functions & 882,657 \\
			...with return type annotations & 94,433 (10.7\%) \\
			\midrule
			Arguments & 1,558,566 \\
			...with type annotations & 128,363 (14.5\%) \\
			\midrule
			Variables & 2,135,361 \\
			...with type annotations & 1,023,328 (47.9\%) \\
			\midrule
			Types & 1,246,124 \\
			...unique & 60,333 \\
			\bottomrule
		\end{tabularx}
		\begin{tablenotes}[flushleft]
			\item[a] {\footnotesize Metrics are counted after the ASTs extraction phase of our pipeline.}
			\item[c] {\footnotesize Comments and blank lines are ignored when counting lines of code.}
		\end{tablenotes}
	\end{threeparttable}
\end{table}

\begin{table}
	\centering
	\caption{Number of datapoints for train, validation and test sets}
	\label{tab:datapoints}
	\begin{tabular}{@{}l l l l@{}}
		\toprule
		& Argument type & Return type & Variable type\\
		\midrule
		Training & 90,114  & 37,803 & 426,235 \\
		Validation & 9,387 & 3,932 & 48,518 \\
		Test & 24,121 & 10,444 & 118,319 \\
		\midrule
		Total &  108,888 (16.06\%) & 45,667 (6.74\%) & 523,271 (77.20\%) \\
		\bottomrule
	\end{tabular}
\end{table}

\subsection{Dataset Characteristics}
Table \ref{tab:dataset} shows the characteristics of our dataset after code de-duplication, augmentation, and type-checking. In total, there are more than 882K functions with around 1.5M arguments. Also, the dataset has more than 2.1M variable declarations. Of which, 48\% have type annotations.

\begin{figure}[!t]
	\centering
	\includegraphics[width=\linewidth]{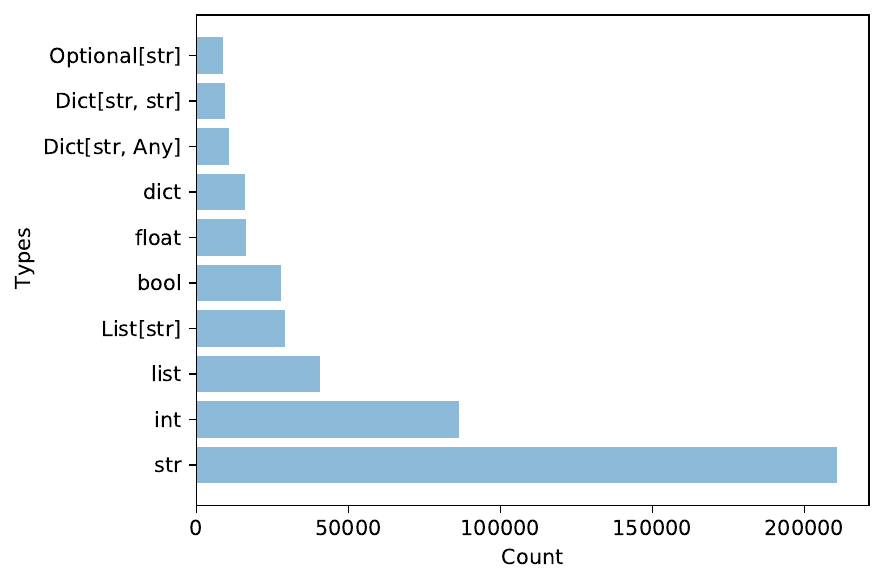}
	\caption{Top 10 most frequent types (\texttt{Any} and \texttt{None} types are excluded)}
	\label{fig:top_10_types}
\end{figure}

Figure \ref{fig:top_10_types} shows the frequency of top 10 most frequent types in our dataset. It can be observed that types follow a long-tail distribution. Unsurprisingly, the top 10 most frequent types amount to 59\% of types in the dataset. Lastly, we randomly split the dataset by files into three sets: 70\% training data, 10\% validation data, and 20\% test data. Table \ref{tab:datapoints} shows the number of data points for each of the three sets.

\subsection{Pre-processing}
Similar to the previous work~\cite{pradel2019typewriter, allamanis2020typilus}, before training ML models, we have performed several pre-processing steps:

\begin{itemize}
	\item Trivial functions such as \code{\_\_str\_\_} and \code{\_\_len\_\_} are not included in the dataset. The return type of this kind of functions is straightforward to predict, i.e., \code{\_\_len\_\_} always returns \code{int}, and would blur the results.
	\item We excluded \code{Any} and \code{None} type annotations as it is not helpful to predict these types.
	\item We performed a simple type aliasing resolving to make type annotations of the same kind consistent. For instance, we map \code{[]} to \code{List}, \code{\{\}} to \code{Dict}, and \code{Text} to \code{str}.
	\item We resolved qualified names for type annotations. For example, \code{array} is resolved to \code{numpy.array}. This makes all the occurrences of a type annotation across the dataset consistent.
	\item Same as the work of Allamanis et al.~\cite{allamanis2020typilus}, we rewrote the components of a base type whose nested level is greater than 2 to \code{Any}. For instance, we rewrite \code{List[List[Tuple[int]]]} to \code{List[List[Any]]]}. This removes very rare types or outliers.
\end{itemize}

\section{Evaluation Setup}\label{sec:es}
In this section, we describe the baseline models, the implementation details and the training of the neural models. Lastly, we explain evaluation metrics to quantitatively measure the performance of ML-based type inference approaches.

\subsection{Baselines}
We compare \name to Typilus~\cite{allamanis2020typilus} and TypeWriter~\cite{pradel2019typewriter}, which are recent state-of-the-art ML-based type inference approaches for Python. Considering Table \ref{tab:comp-learning-appr}, \name has an HNN-based neural model whereas Typilus's neural model is GNN-based. However, Typilus has the same prediction abilities as \name and has no limitation on the size of type vocabulary which makes it an obvious choice for comparison. Compared with \name, TypeWriter has two main differences. First, TypeWriter's type vocabulary is small and pre-defined (i.e. 1,000 types) at training time. Second, TypeWriter cannot predict the type of variables unlike \name and Typilus.

\subsection{Implementation details and environment setup}
\label{sub:implementation}
We implemented \name and TypeWriter in Python 3 and its ecosystem. We extract the discussed type hints from ASTs using LibSA4Py~\cite{libsa4py}. The data processing pipeline is parallelized by employing the \textit{joblib} package. We use NLTK \cite{loper2002nltk} for performing standard NLP tasks such as tokenization and stop work removal. To train the Word2Vec model, the \textit{gensim} package is used. For the neural model, we used bidirectional LSTMs \cite{schuster1997bidirectional} in the PyTorch framework \cite{paszke2019pytorch} to implement the two RNNs. Lastly, we used the Annoy\cite{annoy} package to perform a fast and approximate nearest neighbor search. For Typilus, we used its public implementation on GitHub \cite{typilus}.

We performed all the experiments on a Linux operating system (Ubuntu 18.04.5 LTS).
The computer had an AMD Ryzen Threadripper 1920X with 24 threads (@3.5GHz), 64 GB of RAM, and two NVIDIA GeForce RTX 2080 TIs.

\begin{table}[!t]
	\caption{Value of hyperparameters for neural models}
	\label{tab:hyperp}
	\resizebox{\columnwidth}{!}{\begin{threeparttable}\begin{tabular}{@{}lrrr@{}}
				\toprule
				Hyperparameter &  \name & TypeWriter & Typilus \\
				\midrule
				Word embedding dimension (i.e. $d$) & 100 & 100 & N/A \\
				Size of visible type hints vocabulary (i.e. $T$) & 1024 & 1024 & N/A \\
				LSTM hidden nodes & 256 & 256 & N/A \\
				GNN hidden nodes & N/A & N/A & 64 \\
				Dimension of linear layer's output & 1536 & 1000 & N/A \\
				Number of LSTM's layers & 1 & 1 & N/A \\
				Learning rate & 0.002 & 0.002 & 0.00025 \\
				Dropout rate & 0.25 & 0.25 & 0.1 \\
				Number of epochs & 25 & 25 & 500\tnote{a} \\
				Batch size & 5864 & 4096 & N/A \\
				Value of $k$ for nearest neighbor search & 10 & N/A & 10 \\
				Tripet loss' margin value (i.e. $m$) & 2.0 & N/A & 2.0 \\
				\midrule
				Model's trainable parameters & 4.6M & 4.7M & 650K \\
				\bottomrule
			\end{tabular}
			\begin{tablenotes}
				\item[a] {\footnotesize The model stopped at epoch 38 due to the early stopping technique.}
			\end{tablenotes}
	\end{threeparttable}}
\end{table}

\subsection{Training}
To avoid overfitting the train set, we applied the Dropout regularization \cite{srivastava2014dropout} to the input sequences except for the visible types. Also, we employed the Adam optimizer \cite{kingma2014adam} to minimize the value of the Triplet loss function. For both \name and TypeWriter, we employed the data parallelism feature of PyTorch to distribute training batches between the two GPUs with a total VRAM of 22 GB.  For the \name model, given 554K training samples, a single training epoch takes around 4 minutes. It takes 7 seconds for the TypeWriter model providing that its training set contains 127K training samples\footnote{Note that TypeWriter uses only argument and return samples as it lacks the variable prediction ability.}. Aside from the training sample size, \name is a DSL-based model and hence it has to predict the output of three data points for every single training batch (see Eq. \ref{eq:triplet}). Typilus completes a single training epoch in around 6 minutes\footnote{The public implementation of Typilus does not take advantage of our two GPUs.}. For all the neural models, the validation set is used to find the optimal number of epochs for training. The value of the neural models' hyperparameters is reported in Table \ref{tab:hyperp}.

\subsection{Evaluation Metrics}
We measure the type prediction performance of an approach by comparing the type prediction $t_{p}$ to the ground truth $t_{g}$ using two criteria originally proposed by Allamanis et al.~\cite{allamanis2020typilus}:

\begin{description}
	\item[Exact Match:] $t_{p}$ and $t_{g}$ are exactly the same type.
	\item[Base Type Match:] ignores all type parameters and only matches the base types.
	For example, \code{List[str]} and \code{List[int]} would be considered a match.
\end{description}

In addition to these two criteria, as stated earlier, we opt for the MRR metric~\cite{manning2008introduction}, since the neural models predict a list of types for a given query. The MRR of multiple queries $Q$ is defined as follows:
\begin{equation}
MRR = \frac{1}{|Q|}\sum_{i=1}^{|Q|}{\frac{1}{r_{i}}}
\end{equation}
The MRR metric partially rewards the neural models by giving a score of $\frac{1}{r_{i}}$ to a prediction if the correct type annotation appears in rank $r$. Like Top-1 accuracy, a score of 1 is given to a prediction for which the Top-1 suggested type is correct. Hereafter, we refer to the MRR of the Top-$n$ predictions as MRR@$n$. We evaluate the neural models up to the Top-10 predictions as it is a quite common methodology in the evaluation of ML-based models for code~\cite{ he2021pyart, allamanis2020typilus, pradel2019typewriter}.

Similar to the evaluation methodolgy of Allamanis et al.~\cite{allamanis2020typilus}, we consider types that we have seen more than 100 times in the train set as \emph{common} or \emph{rare} otherwise. Additionally, we define the set of \emph{ubiquitous} types, i.e., $\{\texttt{str}, \texttt{int}, \texttt{list}, \texttt{bool}, \texttt{float}\}$. These types are among the top 10 frequent types in the dataset (see Fig.~\ref{fig:top_10_types}) and they are excluded from the set of common types. Furthermore, Unlike \name and Typilus, TypeWriter predicts \code{unknown} if the expected type is not present in its type vocabulary. Thus, to have a valid comparison with the other two approaches, we consider \code{other} predictions by TypeWriter in the calculation of evaluation metrics.

\section{Evaluation}
\label{sec:eval}
\NewDocumentCommand\RQ{m}{RQ$_{#1}$}

To evaluate and show the effectiveness of \name, we focus on the following research questions.

\begin{description}[noitemsep]
	\item[\RQ{1}] What is the general type prediction performance of \name? 
	\item[\RQ{2}] How does \name perform while considering different predictions tasks?
	\item[\RQ{3}] How do each proposed type hint and the size of type vocabulary contribute to the performance of \name?
\end{description}

\begin{table*}[!t]
	\centering
	\caption{Performance evaluation of the neural models considering different top-$n$ predictions}
	\label{tab:top-n}
	   \begin{threeparttable}
	\begin{tabular}{@{}l l c c c c c c c c c @{}}
		\toprule
		\multirow{2}{*}{Top-$n$ predictions} &  \multirow{2}{*}{Approach} & \multicolumn{4}{c}{\% Exact Match} & & \multicolumn{3}{c}{\% Base Type Match\tnote{a}} & \\
		\cmidrule{3-6} \cmidrule{8-10}
		& & All & Ubiquitous & Common & Rare & & All & Common & Rare & \\
		\midrule
		\multirow{3}{*}{Top-1}  & \name & \textbf{75.8} & \textbf{100.0} & \textbf{82.3} & 19.2 & & \textbf{80.6} & \textbf{85.2} & 36.0 &  \\
		& Typilus  & 66.1 & 92.5 & 73.4 & \textbf{21.6} & & 74.2 & 81.6 & \textbf{41.7} &  \\
		& TypeWriter & 56.1 & 93.5 & 60.9 & 16.2 & & 58.3 & 64.4 & 19.9 & \\
		& & & & & & & & & & \\
		\multirow{3}{*}{Top-3}  & \name & \textbf{78.1} & \textbf{100.0} & \textbf{87.3} & 23.4 & & \textbf{83.8} & \textbf{90.6} & 43.2 &  \\
		& Typilus & 71.6 & 96.2 & 83.0 & \textbf{26.8} & & 79.8 & 88.7 & \textbf{49.2} &  \\
		& TypeWriter  & 63.7 & 98.8 & 79.2 & 20.8 & & 67.3 & 83.5 & 27.9 & \\
		& & & & & & & & & & \\
		\multirow{2}{*}{Top-5}  & \name & \textbf{78.7} & \textbf{100.0} & \textbf{88.6} & 24.5 & & \textbf{84.7} & \textbf{92.1} & 45.5 & \\
		& Typilus & 72.7 & 96.7 & 85.1 & \textbf{28.2} & & 80.9 & 90.1 & \textbf{51.0} & \\
		& TypeWriter & 65.9 & 99.6 & 84.9 & 23.0 & & 70.4  & 89.1 & 32.1 &  \\
		& & & & & & & & & &  \\
		\multirow{2}{*}{Top-10}  &\name & \textbf{79.2} & \textbf{100.0} & 89.7 & 25.2 & & \textbf{85.4} & \textbf{93.3} & 46.9 &   \\
		& Typilus  & 73.3 & 97.04 & 86.4 & \textbf{28.9} & & 81.5 & 90.9 & \textbf{51.9} &  \\
		& TypeWriter & 68.2 & 99.9 & \textbf{90.8} & 25.5 & & 73.2 & 93.8 & 36.5 &  \\
		\midrule
		\multirow{3}{*}{MRR@10} & \name & \textbf{77.1} & \textbf{100.0} & \textbf{85.1} & 21.4 & & \textbf{74.1} & \textbf{79.9} & 29.4 &  \\
		& Typilus  & 69.0 & 94.4 & 78.5 & \textbf{24.4} & & 67.4 & 75.8 & \textbf{32.8} & \\
		& TypeWriter & 60.4 & 96.1 & 71.3 & 19.1 & & 56.5 & 68.0 & 19.7 & \\
		\bottomrule
	\end{tabular}
	\begin{tablenotes}
		\item[a] {\footnotesize Ubiquitous types are not a base type match. However, they are considered in the All column.}
	\end{tablenotes}
\end{threeparttable}
\end{table*}

\begin{table*}
	\centering
	\caption{Performance evaluation of the neural models considering different tasks}
	\label{tab:different-tasks}
	\begin{threeparttable}
		\begin{tabular}{@{}l l l c c c c c c c c c @{}}
			\toprule
			\multirow{2}{*}{Metric} & \multirow{2}{*}{Task} &  \multirow{2}{*}{Approach} & \multicolumn{4}{c}{\% Exact Match} & & \multicolumn{3}{c}{\% Base Type Match}& \\
			\cmidrule{4-7} \cmidrule{9-11}
			& & & All & Ubiquitous & Common & Rare & & All & Common & Rare & \\
			\midrule
			\multirow{10}{*}{Top-1 prediction} & \multirow{3}{*}{Argument} & \name & \textbf{61.9} & \textbf{100.0} & \textbf{64.5} & 17.4 & & \textbf{63.9} & \textbf{69.3} & 20.1 & \\
			& & Typilus  & 53.8 & 83.3 & 46.6 & \textbf{23.7} & & 57.0 & 52.5 & \textbf{29.6} & \\
			& & TypeWriter & 58.4 & 93.6 & 61.3 & 19.6 & & 60.1 & 64.4 & 22.1 & \\
			& & & & & & & & & & & \\
			& \multirow{3}{*}{Return}  & \name & \textbf{56.4} & \textbf{100.0} & 59.3 & \textbf{14.4} & & \textbf{60.3} & \textbf{65.4} & 20.9 & \\
			& & Typilus  & 42.5 & 84.0 & 41.6 & 12.3 & & 49.9 & 49.5 & \textbf{24.8} & \\
			& & TypeWriter  & 50.7 & 93.3 & \textbf{59.9} & 9.2 & & 54.1 & 64.4 & 15.0 & \\
			& & & & & & & & & & & \\
			& \multirow{2}{*}{Variable\tnote{a}}  & \name & \textbf{80.4} & \textbf{100.0} & \textbf{86.8} & 20.7 & & \textbf{85.9} & \textbf{89.1} & 44.6 & \\
			& & Typilus & 71.4 & 95.1 & 80.5 & \textbf{22.5} & & 80.7 & \textbf{89.1} & \textbf{48.6} & \\
			& & & & & & & & & & & \\
			\multirow{10}{*}{MRR@10} & \multirow{3}{*}{Argument} & \name & \textbf{64.2} & \textbf{100.0} & 69.5 & 20.7 & & \textbf{59.9} & 62.2 & 20.6 &  \\
			& & Typilus  & 58.7 & 87.9 & 55.4 & \textbf{27.5} & & 56.0 & 52.2 & \textbf{28.1} & \\
			& & TypeWriter & 63.3 & 96.2 & \textbf{72.4} & 23.0 & & 59.6 & \textbf{69.3} & 22.7 & \\
			& & & & & & & & & & &  \\
			& \multirow{3}{*}{Return}  & \name & \textbf{57.9} & \textbf{100.0} & 63.3 & \textbf{16.1} & & \textbf{52.9} & 55.8 & 18.5 & \\
			& & Typilus  & 46.0 & 86.9 & 49.8 & 14.3 & & 44.9 & 46.6 & \textbf{21.4} & \\
			& & TypeWriter  & 54.2 & 95.9 & \textbf{68.9} & 10.9 & & 49.9 & \textbf{65.1} & 14.2 & \\
			& & & & & & & & & & & \\
			& \multirow{2}{*}{Variable\tnote{a}}  & \name & \textbf{81.4} & \textbf{100.0} & \textbf{89.1} & 22.7 & & \textbf{79.1} & \textbf{85.0} & 34.1 & \\
			& & Typilus & 73.7 & 96.3 & 84.7 & \textbf{25.1} & & 72.4 & 82.7 & \textbf{36.1} & \\
			\bottomrule
		\end{tabular}
		\begin{tablenotes}
			\item[a] {\footnotesize Note that TypeWriter cannot predict the type of variables.}
		\end{tablenotes}
	\end{threeparttable}
\end{table*}

\subsection{Type Prediction Performance (\RQ{1})}
\label{sub:quanti-eval}
In this subsection, we compare our proposed approach, \name, with the selected baseline models in terms of overall type prediction performance.

\paragraph{Method}
The models get trained on the training set and the test set is used to measure the type prediction performance.
We evaluate the neural models by considering different top-$n$ predictions, i.e., $n=\{1,3,5,10\}$.
Also, for this RQ, we consider all the supported inference tasks by the models, i.e., arguments, return types, and variables.

\paragraph{Results}
Table \ref{tab:top-n} shows the overall performance of the neural models while considering different top-$n$ predictions. Given the Top-10 prediction, \name outperforms both Typilus and TypeWriter based on both the exact and base type match criteria (all). Specifically, considering the exact match criteria (all types), \name performs better than Typilus and TypeWriter at the Top-10 prediction by a margin of 5.9\% and 11\%, respectively. Moreover, it can be seen that the \name's performance drop is less significant compared to the other two models when decreasing the value of $n$ from Top-10 to Top-1. For instance, by considering Top-1 rather than Top-10 and the exact match criteria (all), the performance of \name, Typilus, and TypeWriter drop by 3.4\%, 7.2\%, 12.1\%, respectively. Concerning the prediction of rare types, Typilus slightly performs better than \name, which can be attributed to the use of an enhanced triplet loss function. It is also worth mentioning that \name achieves 100\% exact match for the ubiquitous types at Top-1, which is remarkable.

\begin{figure}[!t]
	\centering
	\includegraphics[width=0.75\linewidth]{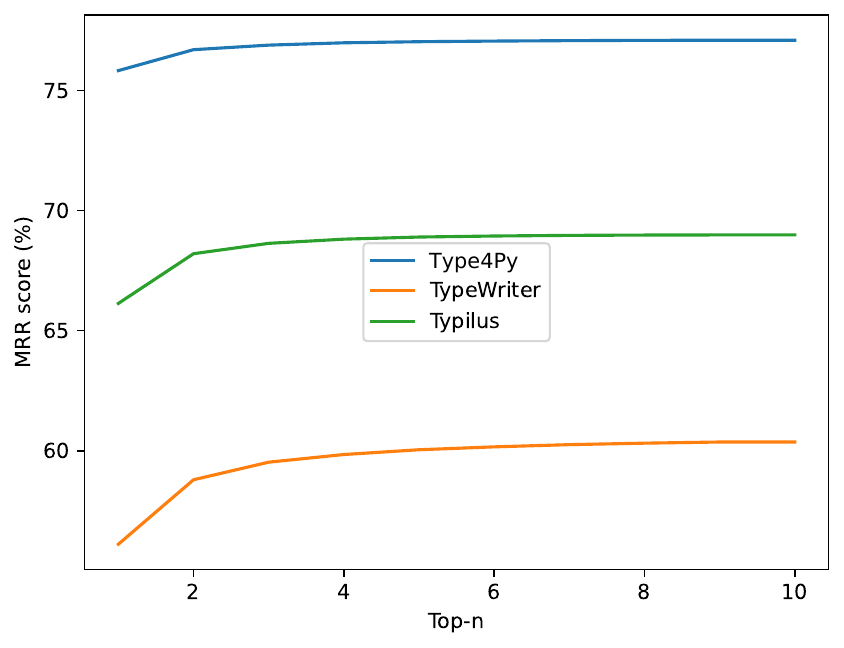}
	\caption{The MRR score of the models considering different top-$n$ predictions}
	\label{fig:top-n-MRR}
\end{figure}

As stated earlier, developers are more likely to use the first suggestion by a tool~\cite{parnin2011automated}. Therefore, we evaluated the neural models by the MRR@10 metric at the bottom of Table \ref{tab:top-n}. Ideally, the difference between the MRR@10 metric and the Top-1 prediction should be zero. However, this is very challenging as the neural models are not 100\% confident in their first suggestion for all test samples. Given the results of MRR@10, we observe that \name outperforms both Typilus and TypeWriter by a margin of 8.1\% and 16.7\%, respectively. In addition, we investigated the MRR score of the neural models while considering different values of Top-$n$, which is shown in Figure \ref{fig:top-n-MRR}. As can be seen, \name has a substantially higher score than the other models across all values of $n$. Moreover, the MRR score of all the three neural models almost converges to a fixed value after MRR@3. Given the findings of the RQ1, we use MRR@10 and the Top-1 prediction for the rest of the evaluation as we believe this better shows the practicality of the neural models for assisting developers.

\begin{table*}[!t]
	\centering
	\caption{Performance evaluation of \name with different configurations}
	\label{tab:models-different-configs}
	\begin{tabular}{@{}l l c c c c c c c c c @{}}
		\toprule
		\multirow{2}{*}{Metric} & \multirow{2}{*}{Approach} & \multicolumn{4}{c}{\% Exact Match} & & \multicolumn{3}{c}{\% Base Type Match\tnote{a}} & \\
		\cmidrule{3-6} \cmidrule{8-10}
		& & All & Ubiquitous & Common & Rare & & All & Common & Rare & \\
		\midrule
		\multirow{5}{*}{Top-1 prediction} & \name & \textbf{75.8} & \textbf{100.0} & 82.3 & \textbf{19.2} & & \textbf{80.6} & 85.2 & \textbf{36.0} & \\		
		& \name (w/o identifiers) & 72.7 & \textbf{100.0} & 71.8 & 17.4 & & 76.5 & 73.9 & 30.9 & \\
		& \name (w/o code context) & 67.9 & \textbf{100.0} & 59.2  & 11.4 & & 70.6 & 63.3 & 17.9 & \\
		& \name (w/o visible type hints)  & 65.4  & 86.2  & 71.9 & 15.8 & & 70.0 & 74.9 & 31.5 & \\
		& \name (w/ top 1,000 types)  & 74.5  & \textbf{100.0}  & \textbf{83.3} & 12.9 & & 79.1 & \textbf{86.3} & 28.5 & \\
		& & & & & & & & & & \\
		\multirow{5}{*}{MRR@10} & \name & \textbf{77.1} & \textbf{100.0} & 85.1  & \textbf{21.4} & & \textbf{74.1} & 79.9 & \textbf{29.4} & \\		
		& \name (w/o identifiers) & 73.8 & \textbf{100.0} & 74.6 & 19.2 & & 69.3 & 66.6 & 25.1 & \\
		& \name (w/o code context) & 69.7 & \textbf{100.0} & 63.9  & 13.6 & & 63.8 & 55.4 & 17.7 & \\
		& \name (w/o visible type hints) & 68.6  & 89.3 & 76.2 & 18.2 & & 65.8 & 70.1 & 26.2 & \\
		& \name (w/ top 1,000 types)  & 75.6  & \textbf{100.0}  & \textbf{86.2} & 14.2 & & 72.4 & \textbf{81.7} & 22.8 & \\
		\bottomrule
	\end{tabular}
\end{table*}

\subsection{Different prediction tasks (\RQ{2})}
Here, we compare \name with other baselines while considering different prediction tasks, i.e., arguments, return types, and variables.

\paragraph{Method}
Similar to the \RQ{1}, the models are trained and tested on the entire training and test sets, respectively.
However, we consider each prediction task separately while evaluating the models at Top-1 and MRR@10.

\paragraph{Results}
Table \ref{tab:different-tasks} shows the type prediction performance of the approaches for the three considered prediction tasks. In general, considering the exact match criteria (all), \name outperforms both Typilus and TypeWriter in all prediction tasks at both Top-1 and MRR@10. For instance, considering the return task and Top-1, \name obtains 56.4\% exact matches (all), which is 13.9\% and 5.7\% higher than that of Typilus and TypeWriter, respectively. Also, for the same task, the \name's MRR@10 is 11.9\% and 3.7\% higher compared to Typilus and TypeWriter, respectively. However, concerning the prediction of common types and MRR@10, TypeWriter performs better than both \name and Typilus at the argument and return tasks. This might be due to the fact that TypeWriter predicts from the set of 1,000 types, which apparently makes it better at the prediction of common types. Moreover, both \name and Typilus have a much larger type vocabulary and hence they need more training samples to generalize better providing that both argument and return types together amount to 22.8\% of all the data points in the dataset (see Table~\ref{tab:datapoints}). Lastly, in comparison with Typilus, \name obtains 7.7\% and 6.7\% higher MRR@10 score for the exact and base type match criteria (all), respectively.

\subsection{Ablation analysis (\RQ{3})}
\label{subsec:abalation}
Here, we investigate how each proposed type hint and the size of type vocabulary contribute to the overall performance of \name.

\paragraph{Method}
For ablation analysis, we trained and evaluated \name with 5 different configurations, i.e., (1) complete model (2) w/o identifiers (3) w/o code context (4) w/o visible type hints (5) w/ a vocabulary of top 1,000 types. Similar to the previous RQs, we measure the performance of \name with the described configurations at Top-1 and MRR@10.

\paragraph{Results}
Table \ref{tab:models-different-configs} presents the performance of \name with the five described configurations. It can be observed that all three type hints contribute significantly to the performance of \name. Code context has the most impact on the model's performance compared to the other two type hints. For instance, when ignoring code context, the model's exact match score for common types drops significantly by 23.1\%. After code context, visible type hints have a large impact on the performance of the model. By ignoring VTH, the model's exact match for ubiquitous types reduces from 100\% to 86.2\%. Although the Identifiers type hint contributes substantially to the prediction of common types, it has a less significant impact on the overall performance of \name compared to code context and VTH. In summary, we conclude that code context and VTH are the strongest type hints for our type prediction model.

By limiting the type vocabulary of \name to top 1,000 types, similar to TypeWriter, we observe that the model's performance for common types is slightly improved while its performance for rare types is reduced significantly, i.e., 7.2\% considering MRR@10. This is expected as the model's type vocabulary is much smaller compared to the complete model's.

\begin{figure*}[!t]
	\centering
	\includegraphics[width=\textwidth]{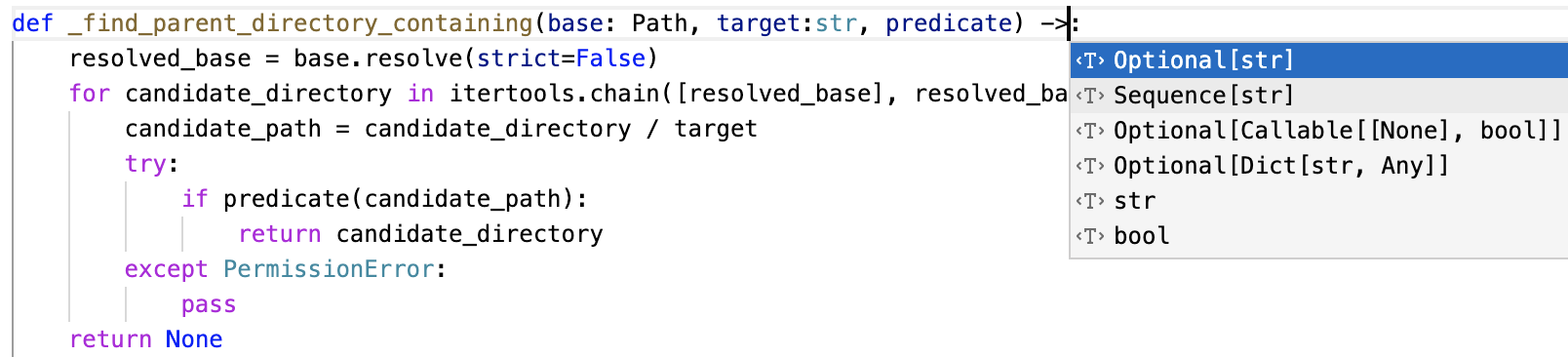}
	\caption{A type auto-completion example from VSC. The code has not seen during training. The expected return type is \code{Optional[str]}.}
	\label{fig:VSC-IDE}
\end{figure*}

\section{Type4Py in Practice}\label{sec:prac}
In this section, we describe the deployment of \name in a production environment, its web server, and Visual Studio Code (VSC) extension

\subsection{Deployment}
To deploy the pre-trained \name model for production, we convert the \name's PyTorch model to an ONNX model \cite{onnxruntime} which enables querying the model on both GPUs and CPUs with faster inference speed. Thanks to Annoy \cite{annoy}, fast and memory-efficient KNN search is performed to suggest type annotations from type clusters.

\subsection{Web server}
We have implemented a small Flask application to handle concurrent type prediction requests from users with Nginx as a proxy. This enables us to have quite a number of asynchronous workers that have an instance of \name’s ONNX model plus Type Clusters each. Specifically, the web application receives a Python source file via a POST request, queries an instance of the model, and finally it gives the file's predicted type annotations as a JSON response.

\subsection{Visual Studio Code extension}
As stated earlier, retrofitting type annotations is a daunting task for developers. To assist developers with this task, we have released a Visual Studio Code extension for \name~\cite{vscodet4py}, which uses the web server's API to provide ML-based type auto-completion for Python code. Figure \ref{fig:VSC-IDE} shows an example of type recommendation from the VSC IDE. As of this writing, the extension has 909 installs on the Visual Studio Marketplace. Based on the user's consent, the VSC extension gathers telemetry data for research purposes. Specifically, accepted types, their rank in the list of suggestions, type slot kind, identifiers' name, and identifiers' line number are captured from the VSC environment and sent to our web server. In addition, rejected type predictions are captured when a type auto-completion window is closed without accepting a type.

By analyzing the gathered telemetry data from Jul. '21 to Aug. '21 and excluding the author(s), of 26 type auto-completion queries, 19 type annotations were accepted by the extension's users. Moreover, the average of accepted type annotations per developer is 69.6\%. Given that the gathered telemetry data is pretty small, we cannot draw a conclusion regarding the performance of \name in practice. However, our telemetry infrastructure and concerted efforts to broaden the user base will enable us to improve \name in the future.

\section{Discussion and future work}\label{sec:discuss}
Based on the formulated RQs and their evaluation in Section \ref{sec:eval}, we provide the following remarks:
\begin{itemize}
	\item We used Pyre~\cite{pyre}, a static type inference tool, to augment our dataset with more type annotations. However, this can be considered as a \emph{weakly} supervision learning problem~\cite{zhou2018brief}, meaning that inferred types by the static tool might be noisy or imprecise despite the pre-processing steps. To eliminate this threat, we employed a static type checker, mypy, to remove source files with type errors from our dataset. Future work can devise a guided-search analysis to fix type errors in source files, which may improve the fix rate.
	
	\item It would be ideal for ML-based models to give a correct prediction in its first few suggestions, preferably Top-1, as developers tend to use the first suggestion by a tool~\cite{parnin2011automated}. Therefore, different from previous work on ML-based type prediction~\cite{pradel2019typewriter, allamanis2020typilus}, we use the MRR metric in our evaluation. We believe that the MRR metric better demonstrates the potential and usefulness of ML models to be used by developers in practice. Overall, considering the MRR metric, \name significantly outperforms the state-the-art ML-based type prediction models, namely, Typilus and TypeWriter.
	
	\item Considering the overall type prediction performance (\RQ{1}), both \name and Typilus generally perform better than TypeWriter. 
	This could be attributed to the fact that the two models map types into a
	high-dimensional space (i.e. type clusters). Hence this not only enables a much
	larger type vocabulary but also significantly improves their overall performance, especially the prediction
	of rare types.
	
	\item Given the results of \RQ{1} and \RQ{2}, our HNN-based neural model, \name, has empirically shown to be more effective than the GNN-based model of Typilus. We attribute this to the inherent bottleneck of GNNs which is over-squashing information into a fixed-size vector \cite{alon2020bottleneck} and thus they fail to capture long-range interaction. However, our HNN-based model concatenates learned features into a high-dimensional vector and hence it preserves information and its long-range dependencies.
	
	\item According to the results of ablation analysis (\RQ{3}), the three
	proposed type hints, i.e., identifiers, code context, and VTHs are
	all effective and positively contribute to the performance of \name. This result does not come
	at the expense of generalizability; our visible type analysis is not more
	sophisticated than what an IDE like PyCharm or VSCode do to determine
	available types for, e.g., auto-completion purposes.
	
	
	\item Both \name and Typilus cannot make a correct prediction for types beyond their
	pre-defined (albeit very large) type clusters. For example, they currently cannot
	synthesize types, meaning that they will never suggest a type such as
	\texttt{Optional[Dict[str, int]]} if it does not exist in their type clusters.
	To address this, future research can explore pointer networks~\cite{vinyals2015pointer} or a GNN model that captures type system rules.
	
	\item We believe that \name's VSC extension is one step forward towards improving developers' productivity by using machine-aided code tools. 
	In this case, the VSC extension aids Python developers to retrofit types for their existing codebases. After gathering sufficiently large telemetry data from the usage of \name, we will study how to improve \name's ranking and quality of predictions for, ultimately, a better user experience.
\end{itemize}

\section{Summary}\label{sec:sum}
In this paper, we present \name, a DSL-based hierarchical neural network type inference model for Python. It considers identifiers, code context, and visible type hints as features for learning to predict types. Specifically, the neural model learns to efficiently map types of the same kind into their own clusters in a high-dimensional space, and given type clusters, the $k$-nearest neighbor search is performed to infer the type of arguments, variables, and functions' return types. We used a type-checked dataset with sound type annotations to train and evaluate the ML-based type inference models. Overall, the results of our quantitative evaluation show that the \name model outperforms other state-of-the-art approaches. Most notably, considering the MRR@10 score, our proposed approach achieves a significantly higher score than that of Typilus and TypeWriter's by a margin of 8.1\% and 16.7\%, respectively. This indicates that our approach gives a more relevant prediction in its first suggestion, i.e., Top-1. Finally, we have deployed \name in an end-to-end fashion to provide ML-based type auto-completion in the VSC IDE and aid developers to retrofit type annotations for their existing codebases.

%
%

\begin{acks}
This research work was funded by H2020 grant 825328 (FASTEN). We thank the anonymous reviewers for their valuable feedback and comments.
\end{acks}

\bibliographystyle{ACM-Reference-Format}
\bibliography{main}


\begin{thebibliography}{66}


\ifx \showCODEN    \undefined \def \showCODEN     #1{\unskip}     \fi
\ifx \showDOI      \undefined \def \showDOI       #1{#1}\fi
\ifx \showISBNx    \undefined \def \showISBNx     #1{\unskip}     \fi
\ifx \showISBNxiii \undefined \def \showISBNxiii  #1{\unskip}     \fi
\ifx \showISSN     \undefined \def \showISSN      #1{\unskip}     \fi
\ifx \showLCCN     \undefined \def \showLCCN      #1{\unskip}     \fi
\ifx \shownote     \undefined \def \shownote      #1{#1}          \fi
\ifx \showarticletitle \undefined \def \showarticletitle #1{#1}   \fi
\ifx \showURL      \undefined \def \showURL       {\relax}        \fi
\providecommand\bibfield[2]{#2}
\providecommand\bibinfo[2]{#2}
\providecommand\natexlab[1]{#1}
\providecommand\showeprint[2][]{arXiv:#2}

\bibitem[\protect\citeauthoryear{Allamanis}{Allamanis}{2019}]%
        {allamanis2018adverse}
\bibfield{author}{\bibinfo{person}{Miltiadis Allamanis}.}
  \bibinfo{year}{2019}\natexlab{}.
\newblock \showarticletitle{The adverse effects of code duplication in machine
  learning models of code}. In \bibinfo{booktitle}{\emph{Proceedings of the
  2019 ACM SIGPLAN International Symposium on New Ideas, New Paradigms, and
  Reflections on Programming and Software}}. \bibinfo{pages}{143--153}.
\newblock


\bibitem[\protect\citeauthoryear{Allamanis, Barr, Ducousso, and Gao}{Allamanis
  et~al\mbox{.}}{2020}]%
        {allamanis2020typilus}
\bibfield{author}{\bibinfo{person}{Miltiadis Allamanis},
  \bibinfo{person}{Earl~T Barr}, \bibinfo{person}{Soline Ducousso}, {and}
  \bibinfo{person}{Zheng Gao}.} \bibinfo{year}{2020}\natexlab{}.
\newblock \showarticletitle{{Typilus: neural type hints}}. In
  \bibinfo{booktitle}{\emph{Proceedings of the 41st ACM SIGPLAN Conference on
  Programming Language Design and Implementation}}. \bibinfo{pages}{91--105}.
\newblock


\bibitem[\protect\citeauthoryear{Alon and Yahav}{Alon and Yahav}{2020}]%
        {alon2020bottleneck}
\bibfield{author}{\bibinfo{person}{Uri Alon} {and} \bibinfo{person}{Eran
  Yahav}.} \bibinfo{year}{2020}\natexlab{}.
\newblock \showarticletitle{On the Bottleneck of Graph Neural Networks and its
  Practical Implications}. In \bibinfo{booktitle}{\emph{International
  Conference on Learning Representations}}.
\newblock


\bibitem[\protect\citeauthoryear{Cheng, Gong, Zhou, Wang, and Zheng}{Cheng
  et~al\mbox{.}}{2016}]%
        {cheng2016person}
\bibfield{author}{\bibinfo{person}{De Cheng}, \bibinfo{person}{Yihong Gong},
  \bibinfo{person}{Sanping Zhou}, \bibinfo{person}{Jinjun Wang}, {and}
  \bibinfo{person}{Nanning Zheng}.} \bibinfo{year}{2016}\natexlab{}.
\newblock \showarticletitle{{Person re-identification by multi-channel
  parts-based cnn with improved triplet loss function}}. In
  \bibinfo{booktitle}{\emph{Proceedings of the iEEE conference on computer
  vision and pattern recognition}}. \bibinfo{pages}{1335--1344}.
\newblock


\bibitem[\protect\citeauthoryear{Chopra, Hadsell, and LeCun}{Chopra
  et~al\mbox{.}}{2005}]%
        {chopra2005learning}
\bibfield{author}{\bibinfo{person}{Sumit Chopra}, \bibinfo{person}{Raia
  Hadsell}, {and} \bibinfo{person}{Yann LeCun}.}
  \bibinfo{year}{2005}\natexlab{}.
\newblock \showarticletitle{{Learning a similarity metric discriminatively,
  with application to face verification}}. In \bibinfo{booktitle}{\emph{2005
  IEEE Computer Society Conference on Computer Vision and Pattern Recognition
  (CVPR'05)}}, Vol.~\bibinfo{volume}{1}. IEEE, \bibinfo{pages}{539--546}.
\newblock


\bibitem[\protect\citeauthoryear{Cover and Hart}{Cover and Hart}{1967}]%
        {cover1967nearest}
\bibfield{author}{\bibinfo{person}{Thomas Cover} {and} \bibinfo{person}{Peter
  Hart}.} \bibinfo{year}{1967}\natexlab{}.
\newblock \showarticletitle{{Nearest neighbor pattern classification}}.
\newblock \bibinfo{journal}{\emph{IEEE transactions on information theory}}
  \bibinfo{volume}{13}, \bibinfo{number}{1} (\bibinfo{year}{1967}),
  \bibinfo{pages}{21--27}.
\newblock


\bibitem[\protect\citeauthoryear{developers}{developers}{2021}]%
        {onnxruntime}
\bibfield{author}{\bibinfo{person}{ONNX~Runtime developers}.}
  \bibinfo{year}{2021}\natexlab{}.
\newblock \bibinfo{title}{{ONNX Runtime}}.
\newblock \bibinfo{howpublished}{\url{https://onnxruntime.ai/}}.
\newblock


\bibitem[\protect\citeauthoryear{Du, Wang, and Wang}{Du et~al\mbox{.}}{2015}]%
        {du2015hierarchical}
\bibfield{author}{\bibinfo{person}{Yong Du}, \bibinfo{person}{Wei Wang}, {and}
  \bibinfo{person}{Liang Wang}.} \bibinfo{year}{2015}\natexlab{}.
\newblock \showarticletitle{{Hierarchical recurrent neural network for skeleton
  based action recognition}}. In \bibinfo{booktitle}{\emph{Proceedings of the
  IEEE conference on computer vision and pattern recognition}}.
  \bibinfo{pages}{1110--1118}.
\newblock


\bibitem[\protect\citeauthoryear{Furr, An, Foster, and Hicks}{Furr
  et~al\mbox{.}}{2009}]%
        {furr2009static}
\bibfield{author}{\bibinfo{person}{Michael Furr}, \bibinfo{person}{Jong-hoon
  An}, \bibinfo{person}{Jeffrey~S Foster}, {and} \bibinfo{person}{Michael
  Hicks}.} \bibinfo{year}{2009}\natexlab{}.
\newblock \showarticletitle{{Static type inference for Ruby}}. In
  \bibinfo{booktitle}{\emph{Proceedings of the 2009 ACM symposium on Applied
  Computing}}. \bibinfo{pages}{1859--1866}.
\newblock


\bibitem[\protect\citeauthoryear{Gao, Bird, and Barr}{Gao
  et~al\mbox{.}}{2017}]%
        {gao2017type}
\bibfield{author}{\bibinfo{person}{Zheng Gao}, \bibinfo{person}{Christian
  Bird}, {and} \bibinfo{person}{Earl~T Barr}.} \bibinfo{year}{2017}\natexlab{}.
\newblock \showarticletitle{{To type or not to type: quantifying detectable
  bugs in JavaScript}}. In \bibinfo{booktitle}{\emph{2017 IEEE/ACM 39th
  International Conference on Software Engineering (ICSE)}}. IEEE,
  \bibinfo{pages}{758--769}.
\newblock


\bibitem[\protect\citeauthoryear{Garcia, de~Carvalho, and Lorena}{Garcia
  et~al\mbox{.}}{2015}]%
        {garcia2015effect}
\bibfield{author}{\bibinfo{person}{Lu{\'\i}s~PF Garcia},
  \bibinfo{person}{Andr{\'e}~CPLF de Carvalho}, {and} \bibinfo{person}{Ana~C
  Lorena}.} \bibinfo{year}{2015}\natexlab{}.
\newblock \showarticletitle{Effect of label noise in the complexity of
  classification problems}.
\newblock \bibinfo{journal}{\emph{Neurocomputing}}  \bibinfo{volume}{160}
  (\bibinfo{year}{2015}), \bibinfo{pages}{108--119}.
\newblock


\bibitem[\protect\citeauthoryear{Goodfellow, Bengio, Courville, and
  Bengio}{Goodfellow et~al\mbox{.}}{2016}]%
        {goodfellow2016deep}
\bibfield{author}{\bibinfo{person}{Ian Goodfellow}, \bibinfo{person}{Yoshua
  Bengio}, \bibinfo{person}{Aaron Courville}, {and} \bibinfo{person}{Yoshua
  Bengio}.} \bibinfo{year}{2016}\natexlab{}.
\newblock \bibinfo{booktitle}{\emph{{Deep learning}}}.
  Vol.~\bibinfo{volume}{1}.
\newblock \bibinfo{publisher}{MIT press Cambridge}.
\newblock


\bibitem[\protect\citeauthoryear{Hanenberg, Kleinschmager, Robbes, Tanter, and
  Stefik}{Hanenberg et~al\mbox{.}}{2014}]%
        {hanenberg2014empirical}
\bibfield{author}{\bibinfo{person}{Stefan Hanenberg},
  \bibinfo{person}{Sebastian Kleinschmager}, \bibinfo{person}{Romain Robbes},
  \bibinfo{person}{{\'E}ric Tanter}, {and} \bibinfo{person}{Andreas Stefik}.}
  \bibinfo{year}{2014}\natexlab{}.
\newblock \showarticletitle{{An empirical study on the impact of static typing
  on software maintainability}}.
\newblock \bibinfo{journal}{\emph{Empirical Software Engineering}}
  \bibinfo{volume}{19}, \bibinfo{number}{5} (\bibinfo{year}{2014}),
  \bibinfo{pages}{1335--1382}.
\newblock


\bibitem[\protect\citeauthoryear{Hassan, Urban, Eilers, and M{\"u}ller}{Hassan
  et~al\mbox{.}}{2018}]%
        {hassan2018maxsmt}
\bibfield{author}{\bibinfo{person}{Mostafa Hassan}, \bibinfo{person}{Caterina
  Urban}, \bibinfo{person}{Marco Eilers}, {and} \bibinfo{person}{Peter
  M{\"u}ller}.} \bibinfo{year}{2018}\natexlab{}.
\newblock \showarticletitle{{Maxsmt-based type inference for python 3}}. In
  \bibinfo{booktitle}{\emph{International Conference on Computer Aided
  Verification}}. Springer, \bibinfo{pages}{12--19}.
\newblock


\bibitem[\protect\citeauthoryear{He, Xu, Zhang, Hao, Feng, and Xu}{He
  et~al\mbox{.}}{2021}]%
        {he2021pyart}
\bibfield{author}{\bibinfo{person}{Xincheng He}, \bibinfo{person}{Lei Xu},
  \bibinfo{person}{Xiangyu Zhang}, \bibinfo{person}{Rui Hao},
  \bibinfo{person}{Yang Feng}, {and} \bibinfo{person}{Baowen Xu}.}
  \bibinfo{year}{2021}\natexlab{}.
\newblock \showarticletitle{PyART: Python API Recommendation in Real-Time}. In
  \bibinfo{booktitle}{\emph{2021 IEEE/ACM 43rd International Conference on
  Software Engineering (ICSE)}}. IEEE, \bibinfo{pages}{1634--1645}.
\newblock


\bibitem[\protect\citeauthoryear{Hellendoorn, Bird, Barr, and
  Allamanis}{Hellendoorn et~al\mbox{.}}{2018}]%
        {hellendoorn2018deep}
\bibfield{author}{\bibinfo{person}{Vincent~J Hellendoorn},
  \bibinfo{person}{Christian Bird}, \bibinfo{person}{Earl~T Barr}, {and}
  \bibinfo{person}{Miltiadis Allamanis}.} \bibinfo{year}{2018}\natexlab{}.
\newblock \showarticletitle{{Deep learning type inference}}. In
  \bibinfo{booktitle}{\emph{Proceedings of the 2018 26th acm joint meeting on
  european software engineering conference and symposium on the foundations of
  software engineering}}. \bibinfo{pages}{152--162}.
\newblock


\bibitem[\protect\citeauthoryear{Hindle, Barr, Su, Gabel, and Devanbu}{Hindle
  et~al\mbox{.}}{2012}]%
        {hindle2012naturalness}
\bibfield{author}{\bibinfo{person}{Abram Hindle}, \bibinfo{person}{Earl~T
  Barr}, \bibinfo{person}{Zhendong Su}, \bibinfo{person}{Mark Gabel}, {and}
  \bibinfo{person}{Premkumar Devanbu}.} \bibinfo{year}{2012}\natexlab{}.
\newblock \showarticletitle{{On the naturalness of software}}. In
  \bibinfo{booktitle}{\emph{2012 34th International Conference on Software
  Engineering (ICSE)}}. IEEE, \bibinfo{pages}{837--847}.
\newblock


\bibitem[\protect\citeauthoryear{Hochreiter and Schmidhuber}{Hochreiter and
  Schmidhuber}{1997}]%
        {hochreiter1997long}
\bibfield{author}{\bibinfo{person}{Sepp Hochreiter} {and}
  \bibinfo{person}{J{\"u}rgen Schmidhuber}.} \bibinfo{year}{1997}\natexlab{}.
\newblock \showarticletitle{{Long short-term memory}}.
\newblock \bibinfo{journal}{\emph{Neural computation}} \bibinfo{volume}{9},
  \bibinfo{number}{8} (\bibinfo{year}{1997}), \bibinfo{pages}{1735--1780}.
\newblock


\bibitem[\protect\citeauthoryear{Jesse, Devanbu, and Ahmed}{Jesse
  et~al\mbox{.}}{2021}]%
        {jesse2021learning}
\bibfield{author}{\bibinfo{person}{Kevin Jesse}, \bibinfo{person}{Premkumar~T
  Devanbu}, {and} \bibinfo{person}{Toufique Ahmed}.}
  \bibinfo{year}{2021}\natexlab{}.
\newblock \showarticletitle{Learning type annotation: is big data enough?}. In
  \bibinfo{booktitle}{\emph{Proceedings of the 29th ACM Joint Meeting on
  European Software Engineering Conference and Symposium on the Foundations of
  Software Engineering}}. \bibinfo{pages}{1483--1486}.
\newblock


\bibitem[\protect\citeauthoryear{Jurafsky and Martin}{Jurafsky and
  Martin}{2009}]%
        {JurafskyNLP}
\bibfield{author}{\bibinfo{person}{Daniel Jurafsky} {and}
  \bibinfo{person}{James~H. Martin}.} \bibinfo{year}{2009}\natexlab{}.
\newblock \bibinfo{booktitle}{\emph{Speech and Language Processing (2nd
  Edition)}}.
\newblock \bibinfo{publisher}{Prentice-Hall, Inc.}, \bibinfo{address}{USA}.
\newblock
\showISBNx{0131873210}


\bibitem[\protect\citeauthoryear{Khan, Chen, Varro, and Mcintosh}{Khan
  et~al\mbox{.}}{2021}]%
        {khan2021empirical}
\bibfield{author}{\bibinfo{person}{Faizan Khan}, \bibinfo{person}{Boqi Chen},
  \bibinfo{person}{Daniel Varro}, {and} \bibinfo{person}{Shane Mcintosh}.}
  \bibinfo{year}{2021}\natexlab{}.
\newblock \showarticletitle{An Empirical Study of Type-Related Defects in
  Python Projects}.
\newblock \bibinfo{journal}{\emph{IEEE Transactions on Software Engineering}}
  (\bibinfo{year}{2021}).
\newblock


\bibitem[\protect\citeauthoryear{Kingma and Ba}{Kingma and Ba}{2014}]%
        {kingma2014adam}
\bibfield{author}{\bibinfo{person}{Diederik~P Kingma} {and}
  \bibinfo{person}{Jimmy Ba}.} \bibinfo{year}{2014}\natexlab{}.
\newblock \showarticletitle{{Adam: A method for stochastic optimization}}.
\newblock \bibinfo{journal}{\emph{arXiv preprint arXiv:1412.6980}}
  (\bibinfo{year}{2014}).
\newblock


\bibitem[\protect\citeauthoryear{Lehtosalo et~al\mbox{.}}{Lehtosalo
  et~al\mbox{.}}{2017}]%
        {lehtosalo2017mypy}
\bibfield{author}{\bibinfo{person}{J Lehtosalo} {et~al\mbox{.}}}
  \bibinfo{year}{2017}\natexlab{}.
\newblock \bibinfo{title}{{Mypy-optional static typing for python}}.
\newblock
\newblock


\bibitem[\protect\citeauthoryear{Liao, Ying~Yang, Zhan, and Rosenhahn}{Liao
  et~al\mbox{.}}{2017}]%
        {liao2017triplet}
\bibfield{author}{\bibinfo{person}{Wentong Liao}, \bibinfo{person}{Michael
  Ying~Yang}, \bibinfo{person}{Ni Zhan}, {and} \bibinfo{person}{Bodo
  Rosenhahn}.} \bibinfo{year}{2017}\natexlab{}.
\newblock \showarticletitle{{Triplet-based deep similarity learning for person
  re-identification}}. In \bibinfo{booktitle}{\emph{Proceedings of the IEEE
  International Conference on Computer Vision Workshops}}.
  \bibinfo{pages}{385--393}.
\newblock


\bibitem[\protect\citeauthoryear{Liu, Zheng, and Zheng}{Liu
  et~al\mbox{.}}{2020}]%
        {liu2020hienn}
\bibfield{author}{\bibinfo{person}{Fagui Liu}, \bibinfo{person}{Lailei Zheng},
  {and} \bibinfo{person}{Jingzhong Zheng}.} \bibinfo{year}{2020}\natexlab{}.
\newblock \showarticletitle{{HieNN-DWE: A hierarchical neural network with
  dynamic word embeddings for document level sentiment classification}}.
\newblock \bibinfo{journal}{\emph{Neurocomputing}}  \bibinfo{volume}{403}
  (\bibinfo{year}{2020}), \bibinfo{pages}{21--32}.
\newblock


\bibitem[\protect\citeauthoryear{Liu, Huang, and Ng}{Liu et~al\mbox{.}}{2018}]%
        {liu2018effective}
\bibfield{author}{\bibinfo{person}{Xiaoyu Liu}, \bibinfo{person}{LiGuo Huang},
  {and} \bibinfo{person}{Vincent Ng}.} \bibinfo{year}{2018}\natexlab{}.
\newblock \showarticletitle{Effective API recommendation without historical
  software repositories}. In \bibinfo{booktitle}{\emph{Proceedings of the 33rd
  ACM/IEEE International Conference on Automated Software Engineering}}.
  \bibinfo{pages}{282--292}.
\newblock


\bibitem[\protect\citeauthoryear{Loper and Bird}{Loper and Bird}{2002}]%
        {loper2002nltk}
\bibfield{author}{\bibinfo{person}{Edward Loper} {and} \bibinfo{person}{Steven
  Bird}.} \bibinfo{year}{2002}\natexlab{}.
\newblock \showarticletitle{{NLTK: The Natural Language Toolkit}}. In
  \bibinfo{booktitle}{\emph{Proceedings of the ACL-02 Workshop on Effective
  Tools and Methodologies for Teaching Natural Language Processing and
  Computational Linguistics}}. \bibinfo{pages}{63--70}.
\newblock


\bibitem[\protect\citeauthoryear{Lopes, Maj, Martins, Saini, Yang, Zitny,
  Sajnani, and Vitek}{Lopes et~al\mbox{.}}{2017}]%
        {lopes2017dejavu}
\bibfield{author}{\bibinfo{person}{Cristina~V Lopes}, \bibinfo{person}{Petr
  Maj}, \bibinfo{person}{Pedro Martins}, \bibinfo{person}{Vaibhav Saini},
  \bibinfo{person}{Di Yang}, \bibinfo{person}{Jakub Zitny},
  \bibinfo{person}{Hitesh Sajnani}, {and} \bibinfo{person}{Jan Vitek}.}
  \bibinfo{year}{2017}\natexlab{}.
\newblock \showarticletitle{{D{\'e}j{\`a}Vu: a map of code duplicates on
  GitHub}}.
\newblock \bibinfo{journal}{\emph{Proceedings of the ACM on Programming
  Languages}} \bibinfo{volume}{1}, \bibinfo{number}{OOPSLA}
  (\bibinfo{year}{2017}), \bibinfo{pages}{1--28}.
\newblock


\bibitem[\protect\citeauthoryear{Madsen}{Madsen}{2015}]%
        {madsen2015static}
\bibfield{author}{\bibinfo{person}{Magnus Madsen}.}
  \bibinfo{year}{2015}\natexlab{}.
\newblock \emph{\bibinfo{title}{{Static analysis of dynamic languages}}}.
\newblock \bibinfo{thesistype}{Ph.D. Dissertation}. \bibinfo{school}{Aarhus
  University}.
\newblock


\bibitem[\protect\citeauthoryear{Maia, Moreira, and Reis}{Maia
  et~al\mbox{.}}{2012}]%
        {maia2012static}
\bibfield{author}{\bibinfo{person}{Eva Maia}, \bibinfo{person}{Nelma Moreira},
  {and} \bibinfo{person}{Rog{\'e}rio Reis}.} \bibinfo{year}{2012}\natexlab{}.
\newblock \showarticletitle{{A static type inference for python}}.
\newblock \bibinfo{journal}{\emph{Proc. of DYLA}} \bibinfo{volume}{5},
  \bibinfo{number}{1} (\bibinfo{year}{2012}), \bibinfo{pages}{1}.
\newblock


\bibitem[\protect\citeauthoryear{Malik, Patra, and Pradel}{Malik
  et~al\mbox{.}}{2019}]%
        {malik2019nl2type}
\bibfield{author}{\bibinfo{person}{Rabee~Sohail Malik}, \bibinfo{person}{Jibesh
  Patra}, {and} \bibinfo{person}{Michael Pradel}.}
  \bibinfo{year}{2019}\natexlab{}.
\newblock \showarticletitle{{NL2Type: inferring JavaScript function types from
  natural language information}}. In \bibinfo{booktitle}{\emph{2019 IEEE/ACM
  41st International Conference on Software Engineering (ICSE)}}. IEEE,
  \bibinfo{pages}{304--315}.
\newblock


\bibitem[\protect\citeauthoryear{Manning, Sch{\"u}tze, and Raghavan}{Manning
  et~al\mbox{.}}{2008}]%
        {manning2008introduction}
\bibfield{author}{\bibinfo{person}{Christopher~D Manning},
  \bibinfo{person}{Hinrich Sch{\"u}tze}, {and} \bibinfo{person}{Prabhakar
  Raghavan}.} \bibinfo{year}{2008}\natexlab{}.
\newblock \bibinfo{booktitle}{\emph{{Introduction to information retrieval}}}.
\newblock \bibinfo{publisher}{Cambridge university press}.
\newblock


\bibitem[\protect\citeauthoryear{Mikolov, Sutskever, Chen, Corrado, and
  Dean}{Mikolov et~al\mbox{.}}{2013}]%
        {mikolov2013distributed}
\bibfield{author}{\bibinfo{person}{Tomas Mikolov}, \bibinfo{person}{Ilya
  Sutskever}, \bibinfo{person}{Kai Chen}, \bibinfo{person}{Greg~S Corrado},
  {and} \bibinfo{person}{Jeff Dean}.} \bibinfo{year}{2013}\natexlab{}.
\newblock \showarticletitle{{Distributed representations of words and phrases
  and their compositionality}}. In \bibinfo{booktitle}{\emph{Advances in neural
  information processing systems}}. \bibinfo{pages}{3111--3119}.
\newblock


\bibitem[\protect\citeauthoryear{Milojkovic, Ghafari, and
  Nierstrasz}{Milojkovic et~al\mbox{.}}{2017}]%
        {milojkovic2017exploiting}
\bibfield{author}{\bibinfo{person}{Nevena Milojkovic},
  \bibinfo{person}{Mohammad Ghafari}, {and} \bibinfo{person}{Oscar
  Nierstrasz}.} \bibinfo{year}{2017}\natexlab{}.
\newblock \showarticletitle{{Exploiting type hints in method argument names to
  improve lightweight type inference}}. In \bibinfo{booktitle}{\emph{2017
  IEEE/ACM 25th International Conference on Program Comprehension (ICPC)}}.
  IEEE, \bibinfo{pages}{77--87}.
\newblock


\bibitem[\protect\citeauthoryear{Mir, Latoskinas, and Gousios}{Mir
  et~al\mbox{.}}{2021}]%
        {mt4py2021}
\bibfield{author}{\bibinfo{person}{Amir~M. Mir}, \bibinfo{person}{Evaldas
  Latoskinas}, {and} \bibinfo{person}{Georgios Gousios}.}
  \bibinfo{year}{2021}\natexlab{}.
\newblock \showarticletitle{ManyTypes4Py: A Benchmark Python Dataset for
  Machine Learning-Based Type Inference}. In \bibinfo{booktitle}{\emph{IEEE/ACM
  18th International Conference on Mining Software Repositories (MSR)}}.
  \bibinfo{publisher}{IEEE Computer Society}, \bibinfo{pages}{585--589}.
\newblock
\urldef\tempurl%
\url{https://doi.org/10.1109/MSR52588.2021.00079}
\showDOI{\tempurl}


\bibitem[\protect\citeauthoryear{[n. d.]}{[n. d.]}{[n.d.]a}]%
        {annoy}
\bibfield{author}{\bibinfo{person}{[n. d.]}.}
  \bibinfo{year}{[n.d.]}\natexlab{a}.
\newblock \bibinfo{title}{{Annoy}}.
\newblock \bibinfo{howpublished}{\url{https://github.com/spotify/annoy}}.
\newblock


\bibitem[\protect\citeauthoryear{[n. d.]}{[n. d.]}{[n.d.]b}]%
        {cd4py}
\bibfield{author}{\bibinfo{person}{[n. d.]}.}
  \bibinfo{year}{[n.d.]}\natexlab{b}.
\newblock \bibinfo{title}{{CD4Py: Code De-Duplication for Python}}.
\newblock \bibinfo{howpublished}{\url{https://github.com/saltudelft/CD4Py}}.
\newblock


\bibitem[\protect\citeauthoryear{[n. d.]}{[n. d.]}{[n.d.]c}]%
        {ieeespec2019}
\bibfield{author}{\bibinfo{person}{[n. d.]}.}
  \bibinfo{year}{[n.d.]}\natexlab{c}.
\newblock \bibinfo{title}{{IEEE Spectrum's the Top Programming Languages
  2021}}.
\newblock
  \bibinfo{howpublished}{\url{https://spectrum.ieee.org/top-programming-languages}}.
\newblock


\bibitem[\protect\citeauthoryear{[n. d.]}{[n. d.]}{[n.d.]d}]%
        {libsa4py}
\bibfield{author}{\bibinfo{person}{[n. d.]}.}
  \bibinfo{year}{[n.d.]}\natexlab{d}.
\newblock \bibinfo{title}{{LibSA4Py: Light-weight static analysis for
  extracting type hints and features}}.
\newblock \bibinfo{howpublished}{\url{https://github.com/saltudelft/libsa4py}}.
\newblock


\bibitem[\protect\citeauthoryear{[n. d.]}{[n. d.]}{[n.d.]e}]%
        {mypy}
\bibfield{author}{\bibinfo{person}{[n. d.]}.}
  \bibinfo{year}{[n.d.]}\natexlab{e}.
\newblock \bibinfo{title}{{Mypy: A static type checker for Python 3}}.
\newblock \bibinfo{howpublished}{\url{https://mypy.readthedocs.io/}}.
\newblock


\bibitem[\protect\citeauthoryear{[n. d.]}{[n. d.]}{[n.d.]f}]%
        {pyre}
\bibfield{author}{\bibinfo{person}{[n. d.]}.}
  \bibinfo{year}{[n.d.]}\natexlab{f}.
\newblock \bibinfo{title}{{Pyre: A performant type-checker for Python 3}}.
\newblock \bibinfo{howpublished}{\url{https://pyre-check.org/}}.
\newblock


\bibitem[\protect\citeauthoryear{[n. d.]}{[n. d.]}{[n.d.]g}]%
        {pyright}
\bibfield{author}{\bibinfo{person}{[n. d.]}.}
  \bibinfo{year}{[n.d.]}\natexlab{g}.
\newblock \bibinfo{title}{{PyRight}}.
\newblock \bibinfo{howpublished}{\url{https://github.com/microsoft/pyright}}.
\newblock


\bibitem[\protect\citeauthoryear{[n. d.]}{[n. d.]}{[n.d.]h}]%
        {pytype}
\bibfield{author}{\bibinfo{person}{[n. d.]}.}
  \bibinfo{year}{[n.d.]}\natexlab{h}.
\newblock \bibinfo{title}{{PyType}}.
\newblock \bibinfo{howpublished}{\url{https://github.com/google/pytype}}.
\newblock


\bibitem[\protect\citeauthoryear{[n. d.]}{[n. d.]}{[n.d.]i}]%
        {vscodet4py}
\bibfield{author}{\bibinfo{person}{[n. d.]}.}
  \bibinfo{year}{[n.d.]}\natexlab{i}.
\newblock \bibinfo{title}{{Type4Py's Visual Studio Code extension}}.
\newblock
  \bibinfo{howpublished}{\url{https://marketplace.visualstudio.com/items?itemName=saltud.type4py}}.
\newblock


\bibitem[\protect\citeauthoryear{[n. d.]}{[n. d.]}{[n.d.]j}]%
        {typilus}
\bibfield{author}{\bibinfo{person}{[n. d.]}.}
  \bibinfo{year}{[n.d.]}\natexlab{j}.
\newblock \bibinfo{title}{{Typilus' public implementation}}.
\newblock \bibinfo{howpublished}{\url{https://github.com/typilus/typilus}}.
\newblock


\bibitem[\protect\citeauthoryear{Ore, Elbaum, Detweiler, and Karkazis}{Ore
  et~al\mbox{.}}{2018}]%
        {ore2018assessing}
\bibfield{author}{\bibinfo{person}{John-Paul Ore}, \bibinfo{person}{Sebastian
  Elbaum}, \bibinfo{person}{Carrick Detweiler}, {and} \bibinfo{person}{Lambros
  Karkazis}.} \bibinfo{year}{2018}\natexlab{}.
\newblock \showarticletitle{{Assessing the type annotation burden}}. In
  \bibinfo{booktitle}{\emph{Proceedings of the 33rd ACM/IEEE International
  Conference on Automated Software Engineering}}. \bibinfo{pages}{190--201}.
\newblock


\bibitem[\protect\citeauthoryear{Pandi, Barr, Gordon, and Sutton}{Pandi
  et~al\mbox{.}}{2020}]%
        {pandi2020opttyper}
\bibfield{author}{\bibinfo{person}{Irene~Vlassi Pandi}, \bibinfo{person}{Earl~T
  Barr}, \bibinfo{person}{Andrew~D Gordon}, {and} \bibinfo{person}{Charles
  Sutton}.} \bibinfo{year}{2020}\natexlab{}.
\newblock \showarticletitle{{OptTyper: Probabilistic Type Inference by
  Optimising Logical and Natural Constraints}}.
\newblock \bibinfo{journal}{\emph{arXiv preprint arXiv:2004.00348}}
  (\bibinfo{year}{2020}).
\newblock


\bibitem[\protect\citeauthoryear{Parnin and Orso}{Parnin and Orso}{2011}]%
        {parnin2011automated}
\bibfield{author}{\bibinfo{person}{Chris Parnin} {and}
  \bibinfo{person}{Alessandro Orso}.} \bibinfo{year}{2011}\natexlab{}.
\newblock \showarticletitle{{Are automated debugging techniques actually
  helping programmers?}}. In \bibinfo{booktitle}{\emph{Proceedings of the 2011
  international symposium on software testing and analysis}}.
  \bibinfo{pages}{199--209}.
\newblock


\bibitem[\protect\citeauthoryear{Paszke, Gross, Massa, Lerer, Bradbury, Chanan,
  Killeen, Lin, Gimelshein, Antiga, et~al\mbox{.}}{Paszke
  et~al\mbox{.}}{2019}]%
        {paszke2019pytorch}
\bibfield{author}{\bibinfo{person}{Adam Paszke}, \bibinfo{person}{Sam Gross},
  \bibinfo{person}{Francisco Massa}, \bibinfo{person}{Adam Lerer},
  \bibinfo{person}{James Bradbury}, \bibinfo{person}{Gregory Chanan},
  \bibinfo{person}{Trevor Killeen}, \bibinfo{person}{Zeming Lin},
  \bibinfo{person}{Natalia Gimelshein}, \bibinfo{person}{Luca Antiga},
  {et~al\mbox{.}}} \bibinfo{year}{2019}\natexlab{}.
\newblock \showarticletitle{{Pytorch: An imperative style, high-performance
  deep learning library}}. In \bibinfo{booktitle}{\emph{Advances in neural
  information processing systems}}. \bibinfo{pages}{8026--8037}.
\newblock


\bibitem[\protect\citeauthoryear{Pavlinovic}{Pavlinovic}{2019}]%
        {pavlinovic2019leveraging}
\bibfield{author}{\bibinfo{person}{Zvonimir Pavlinovic}.}
  \bibinfo{year}{2019}\natexlab{}.
\newblock \emph{\bibinfo{title}{{Leveraging Program Analysis for Type
  Inference}}}.
\newblock \bibinfo{thesistype}{Ph.D. Dissertation}. \bibinfo{school}{New York
  University}.
\newblock


\bibitem[\protect\citeauthoryear{Pradel, Gousios, Liu, and Chandra}{Pradel
  et~al\mbox{.}}{2020}]%
        {pradel2019typewriter}
\bibfield{author}{\bibinfo{person}{Michael Pradel}, \bibinfo{person}{Georgios
  Gousios}, \bibinfo{person}{Jason Liu}, {and} \bibinfo{person}{Satish
  Chandra}.} \bibinfo{year}{2020}\natexlab{}.
\newblock \showarticletitle{{Typewriter: Neural type prediction with
  search-based validation}}. In \bibinfo{booktitle}{\emph{Proceedings of the
  28th ACM Joint Meeting on European Software Engineering Conference and
  Symposium on the Foundations of Software Engineering}}.
  \bibinfo{pages}{209--220}.
\newblock


\bibitem[\protect\citeauthoryear{Rak-amnouykit, McCrevan, Milanova, Hirzel, and
  Dolby}{Rak-amnouykit et~al\mbox{.}}{2020}]%
        {rak2020python}
\bibfield{author}{\bibinfo{person}{Ingkarat Rak-amnouykit},
  \bibinfo{person}{Daniel McCrevan}, \bibinfo{person}{Ana Milanova},
  \bibinfo{person}{Martin Hirzel}, {and} \bibinfo{person}{Julian Dolby}.}
  \bibinfo{year}{2020}\natexlab{}.
\newblock \showarticletitle{Python 3 types in the wild: a tale of two type
  systems}. In \bibinfo{booktitle}{\emph{Proceedings of the 16th ACM SIGPLAN
  International Symposium on Dynamic Languages}}. \bibinfo{pages}{57--70}.
\newblock


\bibitem[\protect\citeauthoryear{Rao, Huang, Feng, and Cong}{Rao
  et~al\mbox{.}}{2018}]%
        {rao2018lstm}
\bibfield{author}{\bibinfo{person}{Guozheng Rao}, \bibinfo{person}{Weihang
  Huang}, \bibinfo{person}{Zhiyong Feng}, {and} \bibinfo{person}{Qiong Cong}.}
  \bibinfo{year}{2018}\natexlab{}.
\newblock \showarticletitle{{LSTM with sentence representations for
  document-level sentiment classification}}.
\newblock \bibinfo{journal}{\emph{Neurocomputing}}  \bibinfo{volume}{308}
  (\bibinfo{year}{2018}), \bibinfo{pages}{49--57}.
\newblock


\bibitem[\protect\citeauthoryear{Ray, Posnett, Filkov, and Devanbu}{Ray
  et~al\mbox{.}}{2014}]%
        {ray2014large}
\bibfield{author}{\bibinfo{person}{Baishakhi Ray}, \bibinfo{person}{Daryl
  Posnett}, \bibinfo{person}{Vladimir Filkov}, {and} \bibinfo{person}{Premkumar
  Devanbu}.} \bibinfo{year}{2014}\natexlab{}.
\newblock \showarticletitle{{A large scale study of programming languages and
  code quality in github}}. In \bibinfo{booktitle}{\emph{Proceedings of the
  22nd ACM SIGSOFT International Symposium on Foundations of Software
  Engineering}}. \bibinfo{pages}{155--165}.
\newblock


\bibitem[\protect\citeauthoryear{Raychev, Vechev, and Krause}{Raychev
  et~al\mbox{.}}{2015}]%
        {raychev2015predicting}
\bibfield{author}{\bibinfo{person}{Veselin Raychev}, \bibinfo{person}{Martin
  Vechev}, {and} \bibinfo{person}{Andreas Krause}.}
  \bibinfo{year}{2015}\natexlab{}.
\newblock \showarticletitle{{Predicting program properties from big code}}. In
  \bibinfo{booktitle}{\emph{ACM SIGPLAN Notices}}, Vol.~\bibinfo{volume}{50}.
  ACM, \bibinfo{pages}{111--124}.
\newblock


\bibitem[\protect\citeauthoryear{Salib}{Salib}{2004}]%
        {salib2004faster}
\bibfield{author}{\bibinfo{person}{Michael Salib}.}
  \bibinfo{year}{2004}\natexlab{}.
\newblock \showarticletitle{{Faster than C: Static type inference with
  Starkiller}}.
\newblock \bibinfo{journal}{\emph{in PyCon Proceedings, Washington DC}}
  (\bibinfo{year}{2004}), \bibinfo{pages}{2--26}.
\newblock


\bibitem[\protect\citeauthoryear{Schuster and Paliwal}{Schuster and
  Paliwal}{1997}]%
        {schuster1997bidirectional}
\bibfield{author}{\bibinfo{person}{Mike Schuster} {and}
  \bibinfo{person}{Kuldip~K Paliwal}.} \bibinfo{year}{1997}\natexlab{}.
\newblock \showarticletitle{{Bidirectional recurrent neural networks}}.
\newblock \bibinfo{journal}{\emph{IEEE transactions on Signal Processing}}
  \bibinfo{volume}{45}, \bibinfo{number}{11} (\bibinfo{year}{1997}),
  \bibinfo{pages}{2673--2681}.
\newblock


\bibitem[\protect\citeauthoryear{Srivastava, Hinton, Krizhevsky, Sutskever, and
  Salakhutdinov}{Srivastava et~al\mbox{.}}{2014}]%
        {srivastava2014dropout}
\bibfield{author}{\bibinfo{person}{Nitish Srivastava},
  \bibinfo{person}{Geoffrey Hinton}, \bibinfo{person}{Alex Krizhevsky},
  \bibinfo{person}{Ilya Sutskever}, {and} \bibinfo{person}{Ruslan
  Salakhutdinov}.} \bibinfo{year}{2014}\natexlab{}.
\newblock \showarticletitle{{Dropout: a simple way to prevent neural networks
  from overfitting}}.
\newblock \bibinfo{journal}{\emph{The journal of machine learning research}}
  \bibinfo{volume}{15}, \bibinfo{number}{1} (\bibinfo{year}{2014}),
  \bibinfo{pages}{1929--1958}.
\newblock


\bibitem[\protect\citeauthoryear{Stuchlik and Hanenberg}{Stuchlik and
  Hanenberg}{2011}]%
        {stuchlik2011static}
\bibfield{author}{\bibinfo{person}{Andreas Stuchlik} {and}
  \bibinfo{person}{Stefan Hanenberg}.} \bibinfo{year}{2011}\natexlab{}.
\newblock \showarticletitle{{Static vs. dynamic type systems: an empirical
  study about the relationship between type casts and development time}}. In
  \bibinfo{booktitle}{\emph{Proceedings of the 7th symposium on Dynamic
  languages}}. \bibinfo{pages}{97--106}.
\newblock


\bibitem[\protect\citeauthoryear{Van~Rossum, Lehtosalo, and Langa}{Van~Rossum
  et~al\mbox{.}}{2014}]%
        {van2014pep}
\bibfield{author}{\bibinfo{person}{Guido Van~Rossum}, \bibinfo{person}{Jukka
  Lehtosalo}, {and} \bibinfo{person}{Lukasz Langa}.}
  \bibinfo{year}{2014}\natexlab{}.
\newblock \showarticletitle{{PEP 484--type hints}}.
\newblock \bibinfo{journal}{\emph{Index of Python Enhancement Proposals}}
  (\bibinfo{year}{2014}).
\newblock


\bibitem[\protect\citeauthoryear{Vinyals, Fortunato, and Jaitly}{Vinyals
  et~al\mbox{.}}{2015}]%
        {vinyals2015pointer}
\bibfield{author}{\bibinfo{person}{Oriol Vinyals}, \bibinfo{person}{Meire
  Fortunato}, {and} \bibinfo{person}{Navdeep Jaitly}.}
  \bibinfo{year}{2015}\natexlab{}.
\newblock \showarticletitle{{Pointer networks}}. In
  \bibinfo{booktitle}{\emph{Advances in neural information processing
  systems}}. \bibinfo{pages}{2692--2700}.
\newblock


\bibitem[\protect\citeauthoryear{Wei, Goyal, Durrett, and Dillig}{Wei
  et~al\mbox{.}}{2019}]%
        {wei2019lambdanet}
\bibfield{author}{\bibinfo{person}{Jiayi Wei}, \bibinfo{person}{Maruth Goyal},
  \bibinfo{person}{Greg Durrett}, {and} \bibinfo{person}{Isil Dillig}.}
  \bibinfo{year}{2019}\natexlab{}.
\newblock \showarticletitle{{LambdaNet: Probabilistic Type Inference using
  Graph Neural Networks}}. In \bibinfo{booktitle}{\emph{International
  Conference on Learning Representations}}.
\newblock


\bibitem[\protect\citeauthoryear{Williams and Zipser}{Williams and
  Zipser}{1989}]%
        {williams1989learning}
\bibfield{author}{\bibinfo{person}{Ronald~J Williams} {and}
  \bibinfo{person}{David Zipser}.} \bibinfo{year}{1989}\natexlab{}.
\newblock \showarticletitle{{A learning algorithm for continually running fully
  recurrent neural networks}}.
\newblock \bibinfo{journal}{\emph{Neural computation}} \bibinfo{volume}{1},
  \bibinfo{number}{2} (\bibinfo{year}{1989}), \bibinfo{pages}{270--280}.
\newblock


\bibitem[\protect\citeauthoryear{Xu, Zhang, Chen, Pei, and Xu}{Xu
  et~al\mbox{.}}{2016}]%
        {xu2016python}
\bibfield{author}{\bibinfo{person}{Zhaogui Xu}, \bibinfo{person}{Xiangyu
  Zhang}, \bibinfo{person}{Lin Chen}, \bibinfo{person}{Kexin Pei}, {and}
  \bibinfo{person}{Baowen Xu}.} \bibinfo{year}{2016}\natexlab{}.
\newblock \showarticletitle{{Python probabilistic type inference with natural
  language support}}. In \bibinfo{booktitle}{\emph{Proceedings of the 2016 24th
  ACM SIGSOFT International Symposium on Foundations of Software Engineering}}.
  ACM, \bibinfo{pages}{607--618}.
\newblock


\bibitem[\protect\citeauthoryear{Zheng, Cai, Chen, Feng, and Chen}{Zheng
  et~al\mbox{.}}{2019}]%
        {zheng2019hierarchical}
\bibfield{author}{\bibinfo{person}{Jianming Zheng}, \bibinfo{person}{Fei Cai},
  \bibinfo{person}{Wanyu Chen}, \bibinfo{person}{Chong Feng}, {and}
  \bibinfo{person}{Honghui Chen}.} \bibinfo{year}{2019}\natexlab{}.
\newblock \showarticletitle{{Hierarchical neural representation for document
  classification}}.
\newblock \bibinfo{journal}{\emph{Cognitive Computation}} \bibinfo{volume}{11},
  \bibinfo{number}{2} (\bibinfo{year}{2019}), \bibinfo{pages}{317--327}.
\newblock


\bibitem[\protect\citeauthoryear{Zhou}{Zhou}{2018}]%
        {zhou2018brief}
\bibfield{author}{\bibinfo{person}{Zhi-Hua Zhou}.}
  \bibinfo{year}{2018}\natexlab{}.
\newblock \showarticletitle{{A brief introduction to weakly supervised
  learning}}.
\newblock \bibinfo{journal}{\emph{National science review}}
  \bibinfo{volume}{5}, \bibinfo{number}{1} (\bibinfo{year}{2018}),
  \bibinfo{pages}{44--53}.
\newblock


\end{thebibliography}

\end{document}